\documentclass{pinepaper}

\usepackage{multirow}
\usepackage{pifont}
\usepackage{algorithm}
\usepackage{algpseudocode}
\usepackage{float}
\usepackage{placeins}
\usepackage{wrapfig}
\newsavebox{\panelcardbox}
\newcommand{\panelcard}[2][\linewidth]{%
  \sbox{\panelcardbox}{\includegraphics[width=#1]{#2}}%
  \begin{tikzpicture}
    \useasboundingbox (0,0) rectangle
      (\the\wd\panelcardbox, \the\ht\panelcardbox);
    \foreach \i in {1,...,7}{%
      \fill[black, opacity=0.028]
        (0.9pt-0.34pt*\i, -1.1pt-0.34pt*\i)
        rectangle
        (\the\wd\panelcardbox+0.9pt+0.34pt*\i,
         \the\ht\panelcardbox-1.1pt+0.34pt*\i);%
    }
    \node[inner sep=0pt, outer sep=0pt, anchor=south west] at (0,0)
      {\usebox{\panelcardbox}};
  \end{tikzpicture}%
}

\definecolor{takeawayfill}{HTML}{FDF3E7}
\definecolor{takeawayline}{HTML}{D98A3D}
\renewenvironment{takeaway}{%
  \begin{tcolorbox}[
    colback=takeawayfill, colframe=takeawayline,
    boxrule=0pt, leftrule=2.4pt, arc=1.2pt, outer arc=1.2pt,
    left=3mm, right=3mm, top=1.8mm, bottom=1.8mm,
    before skip=6pt, after skip=6pt]
  \small\textbf{\sffamily\color{takeawayline}Takeaway.}\ }{\end{tcolorbox}}

\newcommand{\system}{Facet-0}
\newcommand{\dataset}{ManuFacet-1K}

\title{Facet-0: A Robotic Foundation Model for Contact-Rich Precise Manipulation}

\author[1]{Haoyuan Deng\textsuperscript{*}}
\author[1]{Haichao Liu\textsuperscript{*}}
\author[1]{Wenkai Guo\textsuperscript{*}}
\author[1]{Yuan Ling}
\author[1]{Zaijia Yang}
\author[1]{Yuanjiang Xue}
\author[1]{Haosheng Sun}
\author[1]{Liangzi Wang}
\author[1]{Ziwei Wang\textsuperscript{\textdagger}}
\affil[1]{PINE Lab, Nanyang Technological University, Singapore}

\date{\small \textsuperscript{*}Equal contribution. \quad \textsuperscript{\textdagger}Corresponding author.}

\paperlogo{\includegraphics[height=13mm]{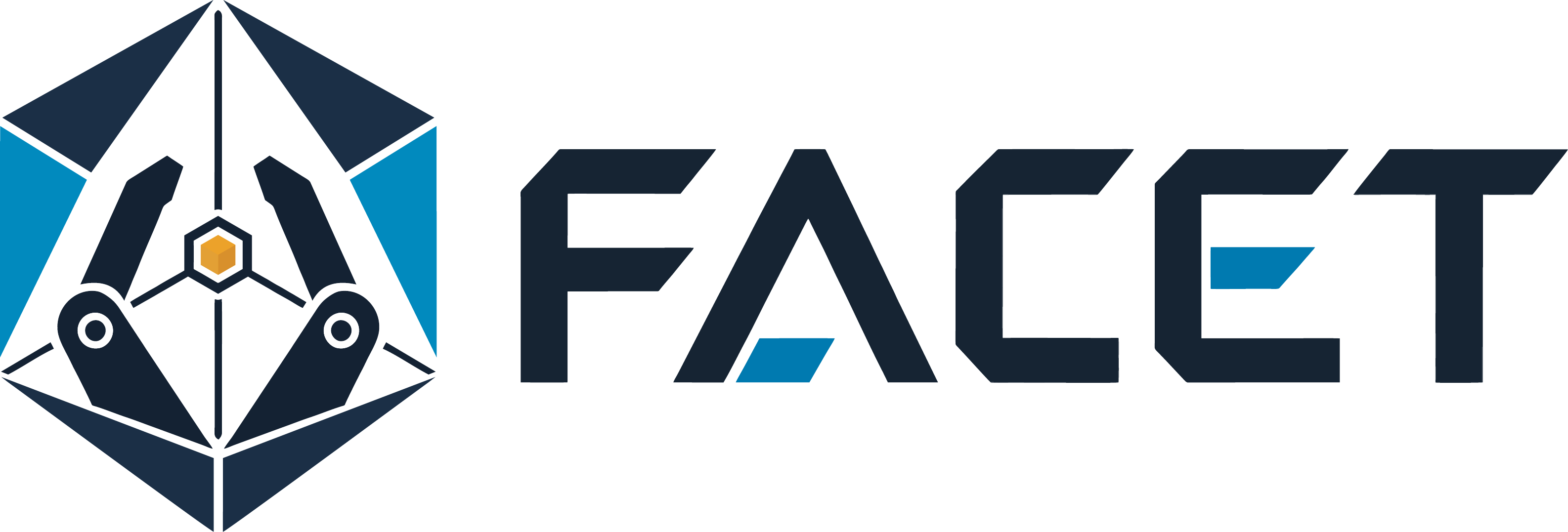}}
\papervenue{Conference / Workshop / Preprint}
\paperdate{\today}
\projectpage{https://pine-lab-ntu.github.io/facet-0/}
\correspondence{\href{mailto:ziwei.wang@ntu.edu.sg}{ziwei.wang@ntu.edu.sg}}

\begin{document}
\makepinetitle

\begin{pineabstract}
Real-world robotic assembly at sub-millimeter tolerances demands spatial precision, compliant interaction, and robustness to contact failures. We present \system{}, a robotic foundation model that predicts and values the contact consequences of its actions. \system{} unifies multimodal representation learning and reinforcement learning (RL) post-training around a joint action--wrench proposal: a causal wrench history is aligned with vision--language semantics and kinematic state, and flow matching generates each action chunk together with the future wrist-wrench profile it is expected to induce. Deployment rollouts train a distributional Action--Wrench Critic to distinguish motions with similar task progress but different contact outcomes, while phase-aware rewards and contact-selective credit concentrate policy improvement on decisive interactions. To accommodate part-specific dynamics, a lightweight bounded actor reuses the frozen representation for on-robot adaptation; RL remains defined over executable Cartesian actions, while an auxiliary wrench head preserves predictive, non-commanded action--contact coupling. Trained on \dataset{}, a 1{,}000-hour force-synchronized corpus spanning three embodiments and multiple manufacturing cells, the bounded task-adapted system reaches 82\% mean success on five sub-millimeter computer-assembly tasks, compared with 15\% for the strongest baseline, with 0.5\,mm placement accuracy and 50\,ms command latency.
\end{pineabstract}


\section{Introduction}
\label{sec:intro}

Precision assembly tests whether robot foundation models can turn broad competence into dependable physical performance. Modern vision--language--action policies follow instructions across tasks and embodiments \cite{black2024pi0,intelligence2025pi05,kim2024openvla}. In manufacturing, recognition and approach bring the robot only to the interface. Success is decided in the final millimeter, where small pose errors become lateral force, friction, or jamming. Visually similar motions can therefore produce opposite physical outcomes. With sub-millimeter clearances and chained operations, one failed contact can invalidate the task \cite{inoue2017peghole,suomalainen2022survey,fmb2024}.

Prior work addresses parts of this gap. Force-aware policies expose interaction signals, while compliant controllers stabilize contact \cite{forcevla2025,tavla2025,tactilevla2025,raibert1981hybrid,hogan1985impedance}. Real-robot reinforcement learning improves behavior from deployment experience \cite{luo2024serl,luo2024hilserl,lei2025rl100}. Yet contact is usually treated as a record of what happened or as low-level feedback, rather than as a future consequence attached to a proposed action and valued by its outcome. We therefore ask: can a generalist robot preserve semantic breadth while anticipating, valuing, and adapting the physical consequences of its actions?

We present \system{} to address this question by treating contact as a predicted consequence of action and as a quantity that acquires value from deployment. Starting from a PaliGemma vision--language backbone with a flow-matching action expert \cite{beyer2024paligemma,lipman2023flow}, the policy aligns multi-view images and the task instruction with kinematic state and a causal history of measured wrist wrench. This produces a semantic--contact representation from which the policy jointly generates a Cartesian action chunk and the future wrist wrench expected after those actions. Only the action is executed, while wrench remains a predictive consequence. The representation is learned on \dataset{}, our 1{,}000-hour force-synchronized corpus spanning three robot embodiments, two chassis families, and multiple manufacturing cells, extending large-scale robot data toward precision contact \cite{oxe2024,droid2024,agibotworld2025}.

Prediction makes contact explicit, but reliable deployment also requires distinguishing useful interaction from costly interaction. Deployment rollouts improve the behavior expressed through the representation by training a distributional Action--Wrench Critic that evaluates each proposed action together with its anticipated wrench \cite{bellemare2017distributional,luo2024hilserl,lei2025rl100}. Motions with similar geometric progress can therefore receive different values when one aligns cleanly and another jams. Contact-selective credit directs this value toward decisive interactions and refines the same generative policy. When part dynamics change, bounded local adaptation reuses the frozen representation to produce executable Cartesian targets while retaining predictive action--contact coupling. Related policy-refinement methods likewise preserve a strong behavior prior while limiting the scope of on-robot updates \cite{wagenmaker2025dsrl,yuan2024policydecorator,residualoffpolicy2025}.

Across five precision computer-assembly tasks, \system{} achieves 82\% mean success, compared with 15\% for the strongest matched RECAP-style baseline, while maintaining 0.5\,mm placement accuracy and 50\,ms command latency \cite{intelligence2025pi05,pi06recap2025,nvidia2026gr00tn17,tavla2025}. Controlled variants obtain 16\% with semantic--contact alignment, 38\% with value-guided refinement, and 82\% with bounded local adaptation. Against advantage-weighted behavior cloning on the same deployment corpus \cite{peng2019advantage}, value-guided learning reduces human intervention from 47\% to 24\% and raises failure recovery from 44\% to 81\%. With ten demonstrations and three hours of training on 6.6\% of the parameters, adaptation reaches 45\% success on an unseen module, compared with 5\% for the strongest baseline. These results connect the final-millimeter problem to measurable gains in completed assemblies.

\begin{figure}[t]
  \centering
  \includegraphics[width=1\linewidth]{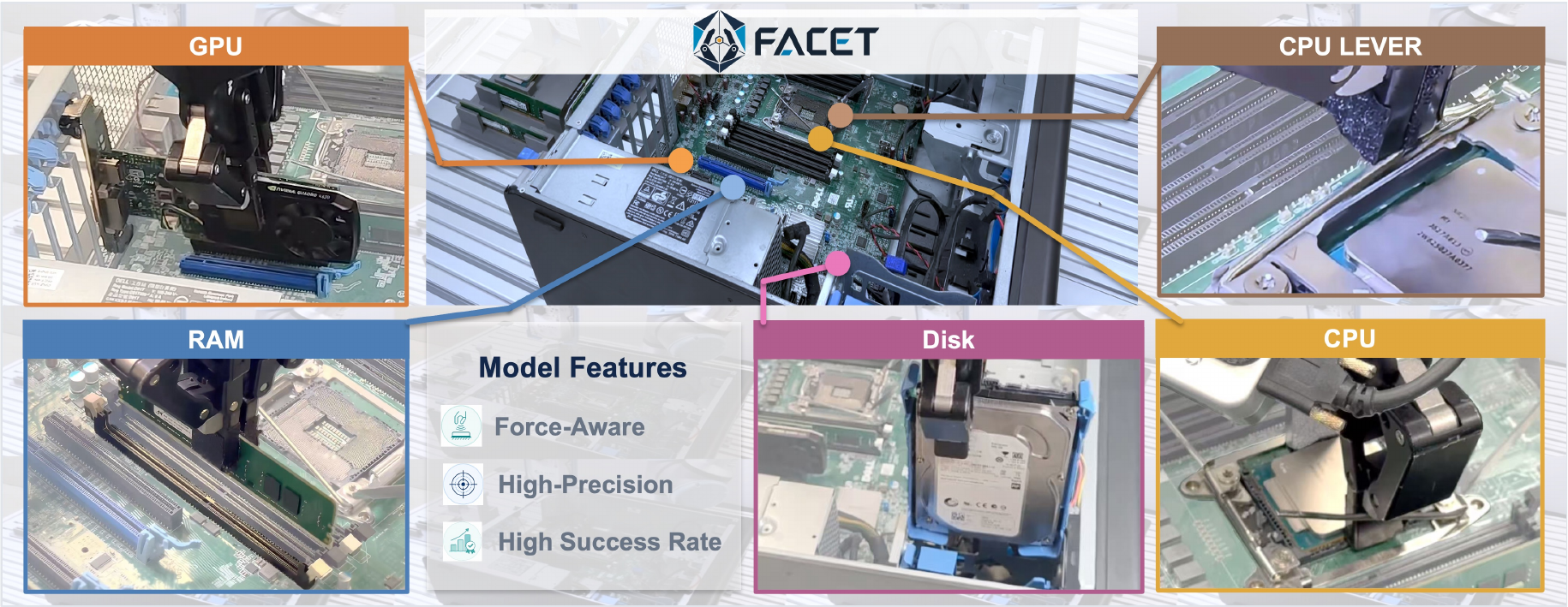}
  \caption{Qualitative overview of \system{} in precision computer assembly. The center shows the robotic workcell, and the surrounding panels show representative GPU, RAM, Disk, CPU, and CPU LEVER operations. Across these contact-rich tasks, \system{} targets precise alignment, force-aware execution, and reliable completion.}
  \label{fig:overview}
\end{figure}

Our contributions are as follows.
\begin{itemize}
  \item \textbf{Force-synchronized manufacturing data.} \dataset{} provides 1{,}000 hours of demonstrations and deployment rollouts across three embodiments and multiple manufacturing cells, complementing broad manipulation corpora with synchronized wrist wrench \cite{oxe2024,droid2024}.
  \item \textbf{A semantic--contact representation.} Joint action--wrench flow matching connects task intent to anticipated physical consequences within a generalist policy \cite{beyer2024paligemma,lipman2023flow}.
  \item \textbf{Value-guided policy improvement.} Deployment value refines decisive contact behavior, while bounded local adaptation reuses the learned representation under changed part dynamics \cite{luo2024hilserl,lei2025rl100}.
  \item \textbf{Precision assembly evidence.} The full model reaches 82\% mean success across five sub-millimeter tasks, with 0.5\,mm accuracy, 50\,ms latency, improved recovery, and transfer to an unseen part against matched generalist and force-aware baselines \cite{intelligence2025pi05,nvidia2026gr00tn17,tavla2025}.
\end{itemize}


\section{Related Work}
\label{sec:related}

\subsection{Vision-Language-Action Foundation Models}
Modern VLA policies combine a pretrained vision--language backbone with an action decoder, spanning autoregressive tokens, diffusion, and flow matching over action chunks \cite{black2024pi0, intelligence2025pi05, kim2024openvla, nvidia2025gr00t, gemini2025robotics, chi2023diffusionpolicy, lipman2023flow}. Knowledge-insulated co-training preserves semantic priors while the action expert learns from robot trajectories \cite{driess2025knowledge}, and efficient action tokenization improves the practical control rate of this family \cite{pertsch2025fast}. These models generalize broadly across scenes and instructions, but their learned output is usually motion alone: contact, when available, is treated as an observation rather than as a consequence attached to the proposed action. That distinction matters in precision assembly, where two motions can reach the same pose while producing a clean insertion or a damaging jam. \system{} retains the semantic breadth of a generalist backbone but changes the object learned by the action expert. It fuses a causal wrench history with visual--language and kinematic context into a semantic--contact representation, then generates action and anticipated wrench jointly. The resulting proposal remains instruction-grounded while exposing interaction quality to the value-guided post-training objective of Sec.~\ref{sec:value_refinement}.

\subsection{Force- and Tactile-Aware Policies}
A rapidly growing body of work integrates contact signals into learned policies. ForceVLA routes six-axis wrench through a force-aware mixture of experts \cite{forcevla2025}. TA-VLA aggregates wrist force/torque history and uses auxiliary wrench prediction to improve the learned interaction representation \cite{tavla2025}. Tactile-VLA couples tactile-aware reasoning with hybrid position--force control \cite{tactilevla2025}, while FoAR uses force feedback in a reactive contact policy \cite{foar2024}. FD-VLA distills a force representation for sensorless inference, and ForceFlow combines force-aware fusion with contact-driven flow matching and force prediction \cite{fdvla2026, forceflow2026}. These results establish the value of contact supervision, but force is usually introduced as an input, auxiliary target, or local control cue within imitation learning. \system{} instead attaches anticipated wrench to the proposed motion and lets deployment value score that joint proposal. Contact therefore influences which interactions the policy improves, not only what it observes.

\subsection{Contact Control and Compliant Assembly}
Classical contact control solved the actuation half of this problem decades ago. Hybrid force-position control partitions the task frame into force-regulated and position-regulated directions \cite{raibert1981hybrid, mason1981compliance}, impedance and admittance control shape the manipulator's apparent dynamics at contact \cite{hogan1985impedance, beltran2020admittance}, variable impedance adapts those dynamics online \cite{abudakka2020vic}, and energy-tank formulations make passivity, and therefore interaction safety, a structural property rather than a tuning outcome \cite{ferraguti2013tank, kronander2016stability}. Peg-in-hole and its relatives have been studied under this apparatus for as long as robots have assembled anything \cite{inoue2017peghole, suomalainen2022survey, villani2016force}. These controllers offer principled contact regulation under their design assumptions, but transferring them across workpieces often requires task-frame identification and parameter retuning. \system{} addresses the complementary learning problem: it anticipates and values contact consequences while leaving compliant control and explicit workspace and per-step motion limits at the execution layer.

\subsection{Learning from Deployment Experience}
Real-world RL has matured into post-training recipes for generalist policies. SERL and HIL-SERL established sample-efficient off-policy learning with human corrections on real robots \cite{luo2024serl, luo2024hilserl}, and a follow-on line attacks the labor those corrections cost, pruning transitions by their influence on policy entropy \cite{deng2026e2hil} or making the intervention itself agentic, triggered from value dynamics rather than by a human takeover \cite{deng2026uniintervene}. RL-100 chains imitation, offline RL and brief online RL into deployment-grade reliability \cite{lei2025rl100}. RECAP scales advantage-conditioned policy extraction to VLA post-training \cite{pi06recap2025}, whereas VLA-RL scales online RL for autoregressive VLAs \cite{lu2025vlarl}; a recent survey maps the space \cite{deng2025rlvlasurvey}. A parallel line adapts frozen policies without touching their weights, steering latent noise \cite{wagenmaker2025dsrl}, learning residual corrections \cite{yuan2024policydecorator, residualoffpolicy2025, silver2018residual, johannink2019residual}, or stabilizing on-robot fine-tuning \cite{chen2025conrft}. We adopt their shared principle, that a learned value estimate rather than the data distribution should select experience, and depart twice: our Action--Wrench Critic scores the predicted pair, so gentle alignment outranks ramming even when both succeed, and the policy is improved in action--wrench space, changing behavior \emph{during} contact rather than only the aim before it. Intervention rate is correspondingly a metric we report, not only a training cost.

\subsection{Manipulation Datasets}
Open X-Embodiment aggregates over one million episodes across embodiments \cite{oxe2024}, DROID contributes 350 hours of diverse scenes \cite{droid2024}, AgiBot World scales to thousands of hours on standardized hardware \cite{agibotworld2025}, and RoboCasa and RoboMIND add simulated and multi-embodiment coverage \cite{robocasa2024, robomind2024}. These resources prioritize behavioral breadth rather than a corpus organized around synchronized six-axis wrench supervision and contact-phase annotation. Assembly-focused resources such as REASSEMBLE and FMB provide contact-rich trajectories or reproducible assembly benchmarks, but at substantially smaller scale \cite{reassemble2025, fmb2024}. \dataset{} complements the first group and scales the second: it is force-synchronized throughout, collected across three embodiments and two chassis families at multiple centers under one hardware and teleoperation specification, and annotated with a skill-phase taxonomy reused for reward decomposition and refinement conditioning.

\section{The \dataset{} Dataset}
\label{sec:dataset}

\dataset{} is an approximately 1{,}000-hour, force-synchronised corpus for precision manipulation. It combines demonstrations and closed-loop rollouts across multiple sites and installation tasks, with a shared state representation and contact-phase vocabulary. This section specifies the corpus and summarises its collection and curation; implementation thresholds and quality-control details are deferred to Appendix~\ref{app:dataset_curation}.

\begin{figure}[t]
  \centering
  \includegraphics[width=\linewidth]{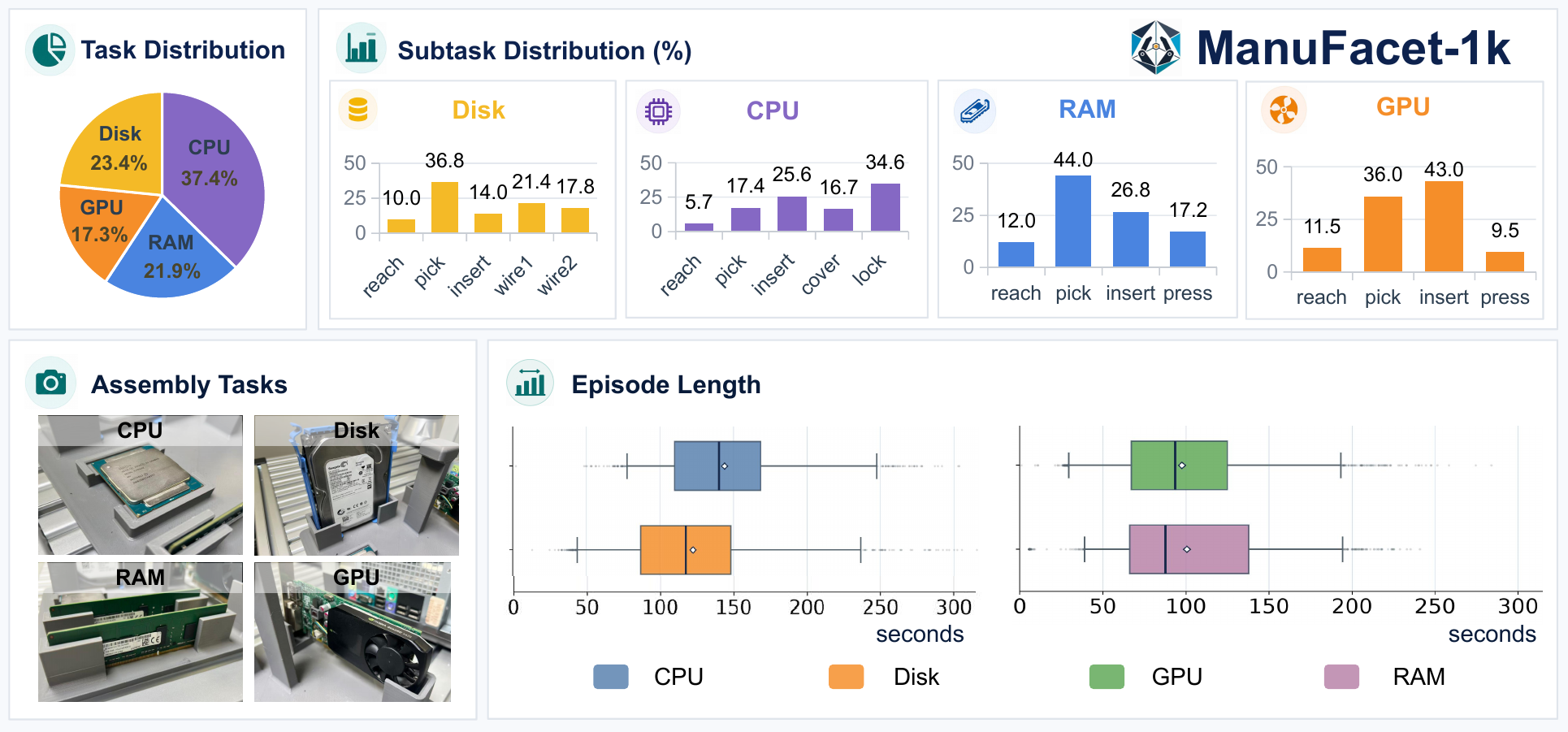}
  \caption{\dataset{} at a glance. The ManuFacet-1K corpus contains approximately 1{,}000 hours of force-synchronized demonstrations and closed-loop rollouts, collected through a multi-center network across four precision installation tasks (CPU 37.3\%, RAM 21.9\%, Disk 23.4\%, GPU 17.3\%), at 0.5\,mm action resolution.}
  \label{fig:dataset}
\end{figure}

\begin{table}[t]
\centering
\caption{Comparison with manipulation corpora using the scale reported by each source. Native units are retained because trajectory counts cannot be converted reliably to hours without duration statistics. ``Mixed'' denotes heterogeneous constituent datasets and ``n/r'' denotes information not reported in the cited source.}
\label{tab:dataset_compare}
\setlength{\tabcolsep}{4pt}
\renewcommand{\arraystretch}{1.12}
\footnotesize
\begin{tabular}{@{}llccc@{}}
\toprule
Dataset & Reported scale & Sync. F/T & Temporal labels & Long-horizon \\
\midrule
Open X-Embodiment \cite{oxe2024} & 1.4M trajectories & mixed & mixed & mixed \\
AgiBot World \cite{agibotworld2025} & 2{,}976.4\,h & n/r & sub-step & \ding{51} \\
RoboCasa \cite{robocasa2024} & 100K+ trajectories & n/r & n/r & \ding{51} \\
RoboMIND \cite{robomind2024} & 305.5\,h & n/r & 10K frame-level & n/r \\
DROID \cite{droid2024} & 350\,h & n/r & n/r & n/r \\
FMB \cite{fmb2024} & $\approx$31\,h & \ding{51} & stage & \ding{51} \\
REASSEMBLE \cite{reassemble2025} & $\approx$13\,h & \ding{51} & action/skill & \ding{51} \\
\rowcolor{gray!10}
\textbf{\dataset{} (ours)} & $\approx$1{,}000\,h & \textbf{\ding{51}} & contact phase & \ding{51} \\
\bottomrule
\end{tabular}
\end{table}

\subsection{Corpus Composition}
The corpus covers four installation tasks in the proportions shown in Fig.~\ref{fig:dataset}: CPU (37.3\%), RAM (21.9\%), Disk (23.4\%), and GPU (17.3\%), spanning diverse contact geometries and force regimes.

Each cell uses a torque-controlled arm with parallel gripper, wrist six-axis force/torque sensing, three RGB cameras, and joint proprioception; kinesthetic teleoperation with force feedback supplies demonstrations in which force regulation is intentional. Every frame stores the 13-D state
\begin{equation}
s_t = \big[\,x_t \in \mathbb{R}^6;\; g_t \in \mathbb{R};\; w_t \in \mathbb{R}^6\,\big],
\label{eq:state}
\end{equation}
combining end-effector pose, gripper opening and wrench.

The curated corpus comprises approximately 1{,}000 hours of open-loop teleoperated demonstrations and closed-loop policy rollouts with human intervention. Action resolution is 0.5\,mm. VLM-generated sub-task/instruction pairs accompany each segment and are reused for pretraining, reward decomposition and adaptation. The release includes a data card, pseudonymised center identities, removed personally identifying content, and curation code.

\subsection{Skill-Phase Taxonomy and Statistics}
We define seven phases over manufacturing skills: \textit{approach}, \textit{align}, \textit{insert}, \textit{press}, \textit{seat}, \textit{fasten}, and \textit{retreat}. The five contact-intended phases (\textit{align}, \textit{insert}, \textit{press}, \textit{seat}, \textit{fasten}) are designated critical because precision determines the outcome. Contact-phase frames are a minority of the corpus but carry most outcome information; peak forces are bimodal, with a low-force alignment mode and a seating mode roughly an order of magnitude higher, depending on the part. Boundaries are marked during collection and verified offline by the VLM pass, with disagreements sent for re-annotation. This shared vocabulary makes phase-indexed frames addressable for pretraining, reward decomposition and action--wrench conditioning. It is distinct from the evaluation skill verbs in Sec.~\ref{sec:exp}: phases describe temporal interaction regimes within a trajectory, whereas verbs name benchmark sub-goals, each of which may traverse multiple phases.

\subsection{Collection and Curation}
Episodes are streamed from a multi-center network under a shared sensing, teleoperation and state specification. Curation consists of five steps: (i) timestamp-based multimodal synchronization; (ii) segment-level validity extraction; (iii) saliency-based action-chunk sampling; (iv) VLM-assisted sub-task and instruction supervision; and (v) anomaly correction with corrected trajectories retained for the data flywheel. Full rates, thresholds, annotator protocol and reflow rules are given in Appendix~\ref{app:dataset_curation}.

\section{Semantic--Contact Alignment}
\label{sec:method}

Precision assembly couples task semantics with contact dynamics that are only partially visible near insertion. Since wrist force/torque reports interaction only after execution, we model each proposed motion jointly with its anticipated wrench, thereby aligning semantic intent with contact consequences.

\begin{figure}[t]
  \centering
  \includegraphics[width=\linewidth]{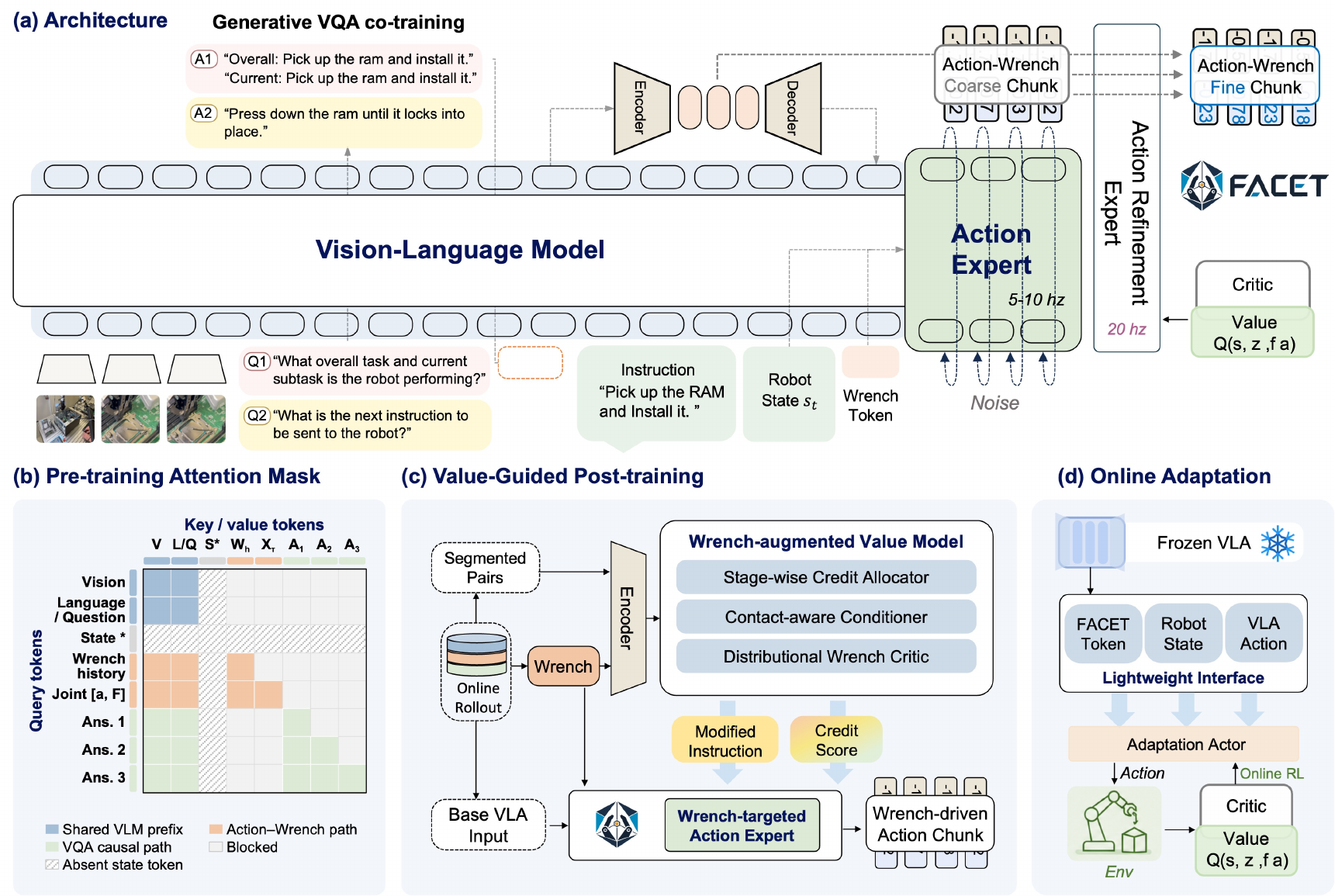}
  \caption{Overview of \system{}. \textbf{(a)} Visual--language context, robot state, and wrench history form a shared semantic--contact representation for coarse-to-fine action--wrench decoding. \textbf{(b)} Structured attention shares the visual--language prefix while separating the action--wrench and causal VQA paths. \textbf{(c)} Phase-segmented rollouts train an Action--Wrench Critic; contact-selective credit then refines the shared action expert. \textbf{(d)} A bounded local actor reuses the frozen representation and policy reference; its critics evaluate executable actions, while its auxiliary wrench prediction remains non-commanded.}
  \label{fig:pipeline}
\end{figure}

\subsection{Multimodal Grounding}
\label{sec:arch}

The base policy pairs a PaliGemma vision--language backbone \cite{beyer2024paligemma} with a flow-matching action expert \cite{lipman2023flow}. At control step $t$, the observation contains three synchronized RGB views $I_t^{1:3}$, a language instruction $\ell$, and the robot state of Eq.~\eqref{eq:state}. We separate that state into the end-effector pose $x_t\in\mathbb R^6$, gripper opening $g_t\in\mathbb R$, and measured wrist wrench $w_t=[f_t;m_t]\in\mathbb R^6$, where $f_t$ and $m_t$ are the three-axis force and moment. The kinematic input is $s_t^{\mathrm{kin}}=[x_t;g_t]\in\mathbb R^7$, while $W_{t-K+1:t}=[w_{t-K+1},\ldots,w_t]\in\mathbb R^{K\times6}$ stores the latest $K$ wrenches. We use $K=10$ so that contact onset, persistent off-axis moment, and rising wrench under stalled motion can be represented as temporal events rather than isolated samples.

The semantic and contact streams are encoded and fused as
\begin{equation}
\begin{aligned}
h_t^{\mathrm{VL}}
  &= E_{\mathrm{VL}}(I_t^{1:3},\ell), \\
h_t^c
  &= F_{\mathrm{fuse}}\!\left(
      h_t^{\mathrm{VL}},
      E_s(s_t^{\mathrm{kin}}),
      E_w(W_{t-K+1:t})\right),
\end{aligned}
\label{eq:fusion}
\end{equation}
Here $E_{\mathrm{VL}}$, $E_s$, and $E_w$ are respectively the vision--language, kinematic-state, and wrench-history encoders; $F_{\mathrm{fuse}}$ is the learned cross-modal fusion map; and $h_t^c$ is the semantic--contact representation. Conditioned on $h_t^c$, the flow-matching policy $\pi_\theta$ with parameters $\theta$ samples a horizon-$H$ joint proposal $Y_t\sim\pi_\theta(\cdot\mid h_t^c)$. We write this proposal as $Y_t=[A_{t:t+H-1},\widehat W_{t+1:t+H}]\in\mathbb R^{H\times13}$, where $A_{t:t+H-1}\in\mathbb R^{H\times7}$ contains Cartesian-pose and gripper commands and $\widehat W_{t+1:t+H}\in\mathbb R^{H\times6}$ contains the wrist wrenches predicted after those commands, expressed in physical units after reversing the training normalization. Thus row $k$ pairs $a_{t+k}$ with its anticipated consequence $\widehat w_{t+k+1}$. We use $H=50$. Only $A$ is executable; $\widehat W$ is an anticipated contact consequence, not a force command.

As shown in Fig.~\ref{fig:pipeline}(a), the action expert produces a coarse action--wrench chunk at $5$--$10$\,Hz and the refinement expert updates it at $20$\,Hz without changing the joint parameterization. The structured mask in Fig.~\ref{fig:pipeline}(b) aligns the two learning paths without target leakage: visual and language/question tokens form their shared prefix; the action--wrench path additionally reads $W_h=E_w(W_{t-K+1:t})$ and the noised joint chunk $X_\tau$ defined below, whereas the VQA path attends causally to preceding answer tokens. Cross-attention between their targets is blocked, and robot state is injected only into the action expert, as indicated by the hatched state row and column.

\subsection{Predictive Interaction Learning}
\label{sec:predictive_interaction}

Each demonstration provides a clean data target $Y_t^{\mathrm{data}}=[A_{t:t+H-1}^{\mathrm{data}},W_{t+1:t+H}^{\mathrm{data}}]\in\mathbb R^{H\times13}$, whose columns contain executed actions and the measured wrenches that follow them. Conditional flow matching \cite{lipman2023flow} interpolates this target with Gaussian noise $\epsilon\sim\mathcal N(0,\mathbf I)$ of the same shape, where $\mathbf I$ is the identity covariance. For a sampled interpolation time $\tau\in[0,1]$, let $X_\tau=(1-\tau)Y_t^{\mathrm{data}}+\tau\epsilon$ and $U_\tau=\epsilon-Y_t^{\mathrm{data}}$. The learned velocity field $v_\theta$ is optimized by
\begin{equation}
\begin{aligned}
\mathcal L_{\mathrm{AW}}(\theta)
  ={}& \mathbb E\!\left[
      \left\|v_\theta^a(X_\tau,\tau\mid h_t^c)-U_\tau^a\right\|_2^2
      \right] \\
  &+\lambda_{\mathrm{pre}}\mathbb E\!\left[
      \left\|v_\theta^w(X_\tau,\tau\mid h_t^c)-U_\tau^w\right\|_2^2
      \right].
\end{aligned}
\label{eq:awloss}
\end{equation}
The superscripts $a$ and $w$ select the seven action and six wrench columns, respectively; validity normalization is applied independently to the available entries in each slice. The scalar $\lambda_{\mathrm{pre}}$ balances wrench prediction against action learning, and the expectation is over demonstrations, $\tau$, and $\epsilon$. At inference, integrating $v_\theta$ backward from noise at $\tau=1$ to data at $\tau=0$ produces the joint proposal $Y_t$ defined above; consequently, lowering the wrench error requires the policy to encode how its proposed motion changes contact.

The semantic objective $\mathcal L_{\mathrm{VQA}}$ is answer-token cross-entropy for two questions: the overall/current sub-task and the next instruction. Separate optimizers alternate between $\mathcal L_{\mathrm{AW}}$ and $\mathcal L_{\mathrm{VQA}}$ rather than mixing their heterogeneous scales. Combined with the structured mask, this trains a shared semantic prefix while preserving distinct causal paths for interaction and language supervision. Held-out optimization diagnostics are reported in Appendix~\ref{app:alignment_diagnostics}.

\section{Value-Guided Policy Refinement}
\label{sec:value_refinement}

The aligned policy proposes plausible interactions but cannot rank them by task utility.
Deployment rollouts supply that ranking through sub-goal completion, contact quality, and recovery from failure.
Within the aligned policy, the coarse expert supplies a horizon-scale joint proposal and the refinement expert updates that same action--wrench object near contact.
We learn a value function over these proposals and use it to refine the shared generative policy.
For part-specific transfer, a downstream local actor instead consumes the frozen representation and policy reference; its critic acts only on the seven-dimensional executable action and does not alter the global proposal space.

\begin{figure}[t]
  \centering
  \includegraphics[width=0.95\linewidth]{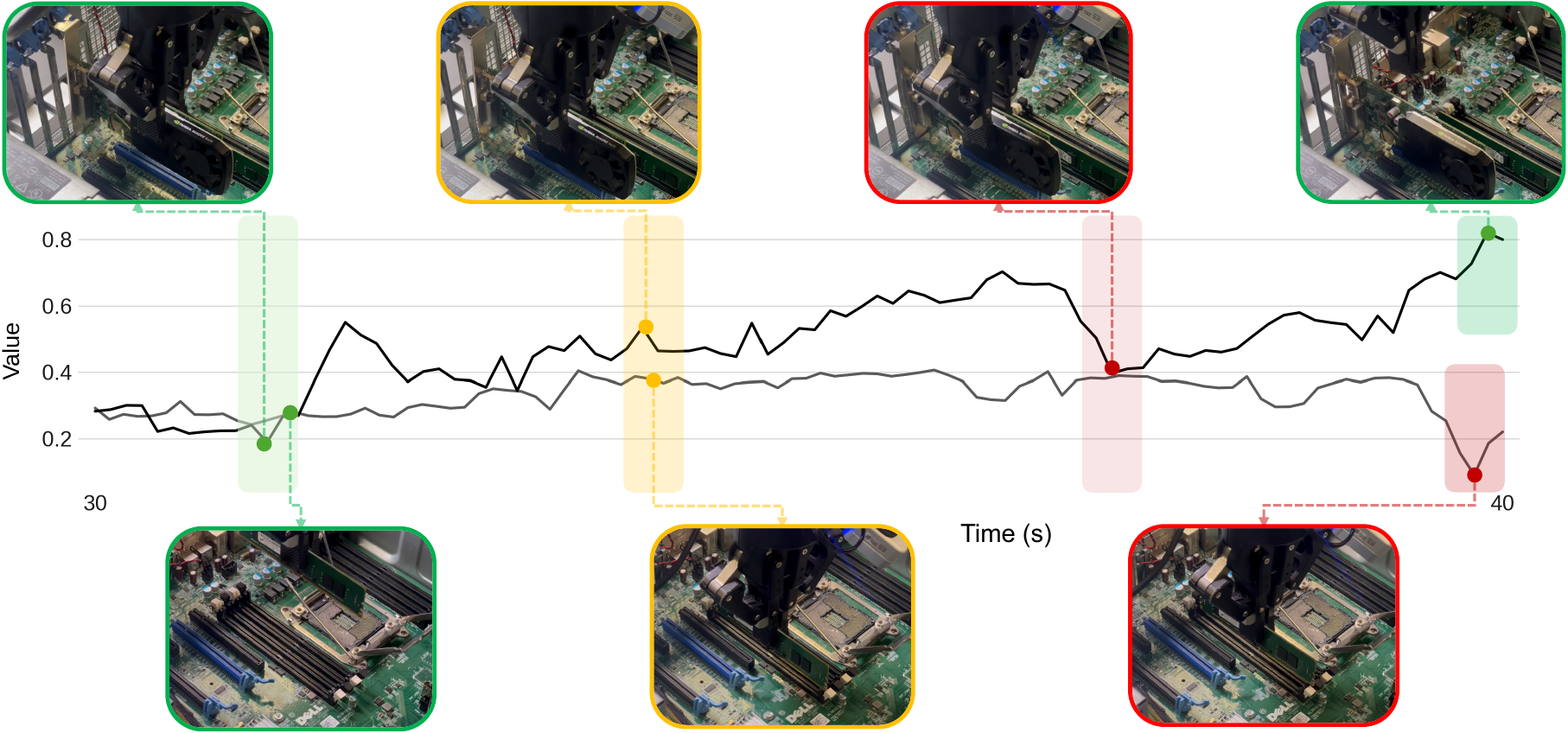}
  \caption{Action--Wrench Critic values during deployment. A value estimate based on vision and state alone remains insensitive to fine-grained contact, whereas scoring the predicted action--wrench proposal separates a clean GPU insertion from a failure mode that is visually difficult to distinguish.}
  \label{fig:value_vis}
\end{figure}

\subsection{Interaction Valuation}
\label{sec:interaction_value}

We first utilize the ManuFacet-1k dataset to train a VLM to provide a vision--language judge $\Omega$ that segments each rollout into sub-goals, marking completion with $\Omega_t\in\{0,1\}$ and labeling the contact phase $\varphi_t$.
Let $\lVert w_t\rVert_{\varphi_t}$ denote the six-axis wrench normalized per axis by the envelope configured for phase $\varphi_t$, so any value above one is a violation, and let $\Delta t$ be the time spent in the current sub-goal against its budget $T_{\max}$.
Progress and contact then share a single per-step signal,
\begin{equation}
r_t=\Omega_t\Big(1+\beta\big[1-\tfrac{\Delta t}{T_{\max}}\big]_+\Big)-\alpha\big[\lVert w_t\rVert_{\varphi_t}-1\big]_+,
\label{eq:reward}
\end{equation}
where $[x]_+=\max(x,0)$. Thus completing a sub-goal earns credit, completing it sooner earns more, and forcing through it earns less; $\alpha,\beta\ge0$ set the exchange rate.

Using the notation of Sec.~\ref{sec:method}, $Y_t\in\mathbb R^{H\times13}$ is the joint action--wrench proposal and $h_t^c$ is the semantic--contact representation.
The critic $Z_\psi(h_t^c,Y_t)$ predicts the return distribution of executing the action component of $Y_t$ \cite{bellemare2017distributional}; its mean $Q_\psi=\mathbb E[Z_\psi]$ is the Action--Wrench Critic, and $V_\psi(h_t^c)=\mathbb E_{Y\sim\pi_\theta}[Q_\psi(h_t^c,Y)]$ is the value of the policy's own proposal.
Training is temporal difference on stored rollouts,
\begin{equation}
\mathcal L_Q(\psi)=\mathbb E\Big[\mathcal D_{\mathrm{dist}}\big(Z_\psi(h_t^c,Y_t),\;r_t+\gamma(1-d_t)Z_{\bar\psi}(h_{t+1}^c,Y_{t+1})\big)\Big],
\label{eq:critic}
\end{equation}
where $\mathcal D_{\mathrm{dist}}$ is the distributional regression loss, $\gamma$ is the discount, $d_t\in\{0,1\}$ marks a terminal transition, $\bar\psi$ denotes target parameters, and $Y_{t+1}\sim\pi_\theta(\cdot\mid h_{t+1}^c)$ is the successor proposal. The factor $1-d_t$ removes the bootstrap at terminal transitions.

Two properties of this critic matter downstream.
Reading the predicted wrench alongside the action lets it separate a clean insertion from a jam stalled at the same depth, which are indistinguishable by geometric progress alone, as visualized in Fig.~\ref{fig:value_vis}.
Predicting a distribution rather than a mean preserves the success and failure modes that contact-rich outcomes split into, instead of averaging them into a value no frame actually attains.
Rollouts are stored as adjacent, phase-segmented transitions and replayed offline, shown as ``Segmented Pairs'' in Fig.~\ref{fig:pipeline}(c).

\subsection{Contact-Selective Credit}

Ranking frames by return favors those that make visible progress, and decisive contact is not among them: it occupies few frames and advances measured progress slowly.
We therefore score each frame by a short-horizon, undiscounted credit
\begin{equation}
\delta_t^{(N)}=\sum_{j=0}^{N-1}r_{t+j}+V_\psi(h_{t+N}^c)-V_\psi(h_t^c),\qquad N<H,
\label{eq:advantage}
\end{equation}
which measures how far a frame outruns the progress rate the critic already expects at that point.
Here $N$ is the credit horizon, chosen shorter than the action-chunk horizon $H$. Being undiscounted, $\delta_t^{(N)}$ is a within-rollout credit score rather than a Bellman-consistent $N$-step advantage, which is all the label below requires.

A single global threshold on $\delta_t^{(N)}$ would still select mostly free-space frames.
We instead split frames by contact regime $\rho_t\in\{\mathrm{contact},\mathrm{free}\}$ and take the top-ranked fraction within each. If $q_{\rho_t}$ is the within-regime score threshold, the positive-interaction label is $b_t=\mathbf 1[\delta_t^{(N)}\ge q_{\rho_t}]$, where $\mathbf 1[\cdot]$ is the indicator function.
The normalized contact intensity $c_t\in[0,1]$ then weights the refinement loss,
\begin{equation}
\mathcal L_{\mathrm{post}}(\theta)=\mathbb E_t\big[(1+\lambda c_t)\,\mathcal L_{\mathrm{FM}}(\theta;Y_t^{\mathrm{data}},h_t^c,b_t)\big],
\label{eq:rpt}
\end{equation}
where $\mathcal L_{\mathrm{FM}}$ is the per-sample action--wrench flow-matching loss of Eq.~\eqref{eq:awloss}, $\lambda\ge0$ sets the contact upweighting, and $Y_t^{\mathrm{data}}$ is the executed-action/realized-wrench target defined in Sec.~\ref{sec:predictive_interaction}. In contrast, $Y_t$ without a superscript always denotes the policy's predicted joint proposal scored by the critic.

The label reaches the policy as a short tag appended to the instruction: positive frames train the guided branch, and the unconditional branch trains on all frames with the tag dropped.
This is the contact-selective credit mechanism of Fig.~\ref{fig:pipeline}(c), where $\delta_t^{(N)}$ is the displayed credit score, $b_t$ is delivered as a tagged instruction to the shared action expert, and $Q_\psi$ supplies the value that guides the $20$\,Hz refinement expert.

\subsection{Local Policy Adaptation}
\label{sec:local_policy_adaptation}

Part changes primarily alter local contact dynamics while leaving task semantics and coarse geometry intact. The frozen bottleneck encoder $E_z$ compresses the semantic--contact representation $h_t^c$ learned in Sec.~\ref{sec:method} into the FACET token $z_t=E_z(h_t^c)$. The frozen policy also supplies a reference action $\widetilde a_t\in\mathbb R^7$. Learned projections followed by layer normalization map $z_t$, the kinematic state $[x_t;g_t]$, the measured wrench $w_t$, and $\widetilde a_t$ to modality embeddings $e_t^z$, $e_t^p$, $e_t^w$, and $e_t^{\mathrm{ref}}$, respectively. Their concatenation $e_t=[e_t^z;e_t^p;e_t^w;e_t^{\mathrm{ref}}]$ is the lightweight actor interface in Fig.~\ref{fig:pipeline}(d), with dimensions detailed in Appendix~\ref{app:local_policy_adaptation}.

Let $f_\eta$ be the shared actor with parameters $\eta$. Its two heads return a raw action activation $u_t^a\in\mathbb R^7$ and the next-wrench prediction $\widehat w_{t+1}\in\mathbb R^6$, written $f_\eta(e_t)=[u_t^a,\widehat w_{t+1}]$. A componentwise $\tanh$ followed by affine rescaling maps $u_t^a$ into the task-specific box $[a_{\min},a_{\max}]$, where $a_{\min},a_{\max}\in\mathbb R^7$, to obtain the bounded absolute Cartesian target $a_t\in\mathbb R^7$. The existing safety interface applies workspace and per-step motion limits immediately before execution. The actor is trained with
\begin{equation}
\begin{aligned}
\mathcal L_{\mathrm{actor}}(\eta)
  &= -\mathbb E\!\left[Q_{\xi_1}
       (e_t,a_t)\right]
     +\lambda_{\mathrm{BC}}\mathcal L_a
       (a_t,a_t^{\mathrm{human}}) \\
  &\hspace{2.3em}
     +\lambda_{\mathrm{pred}}\mathcal L_w
       (\widehat w_{t+1},w_{t+1}).
\end{aligned}
\label{eq:local_actor_objective}
\end{equation}
The expectation is over online actor states. The first term improves the actor under one of the twin TD3 critics \cite{fujimoto2018td3}, denoted $Q_{\xi_i}$ for $i\in\{1,2\}$ with parameters $\xi_i$; both critics have domain $(e_t,a_t)$ and evaluate only the executable seven-dimensional action. Actor updates query the bounded proposal, whereas critic regression uses the safety-filtered command stored with each transition. In the remaining terms, $a_t^{\mathrm{human}}$ is a successful intervention action, $\mathcal L_a$ is its action-anchor loss, and $\mathcal L_w$ compares the predicted wrench with the measured next-step target $w_{t+1}$; $\lambda_{\mathrm{BC}}$ and $\lambda_{\mathrm{pred}}$ are their respective weights.

The frozen reference $\widetilde a_t$ is a conditioning prior, not a residual base. The predicted wrench is fully implemented and supervised as an auxiliary contact-consequence target, but is neither a TD3 action dimension nor a force command. Accordingly, the local critic is shown in Fig.~\ref{fig:pipeline}(d) as $Q(e,a)$: it evaluates the actor interface and executable action, not the auxiliary prediction. At deployment only $a_t$ is executed. The wrench head remains active but non-commanded, and compliant control closes its force loop using measured wrist F/T feedback.

\section{Experiments}
\label{sec:exp}

We evaluate \system{} as a deployed system rather than as a policy in isolation. The experiments answer four questions. (Q1) Does the full model beat generalist and force-aware VLAs on precision assembly, and how do semantic--contact alignment, value-guided refinement, and local adaptation contribute (Sec.~\ref{sec:exp_main})? (Q2) What does post-training from deployment experience buy that supervised post-training does not (Sec.~\ref{sec:exp_posttrain})? (Q3) How does the local actor compare against other adaptation mechanisms, and how cheaply does it transfer to an unseen part (Sec.~\ref{sec:exp_adapt})? (Q4) Which uses of the wrench signal carry the gains (Sec.~\ref{sec:exp_ablation})?

\subsection{Experimental Setup}
\label{sec:exp_setup}

\begin{figure}[t]
  \centering
  \includegraphics[width=0.98\linewidth]{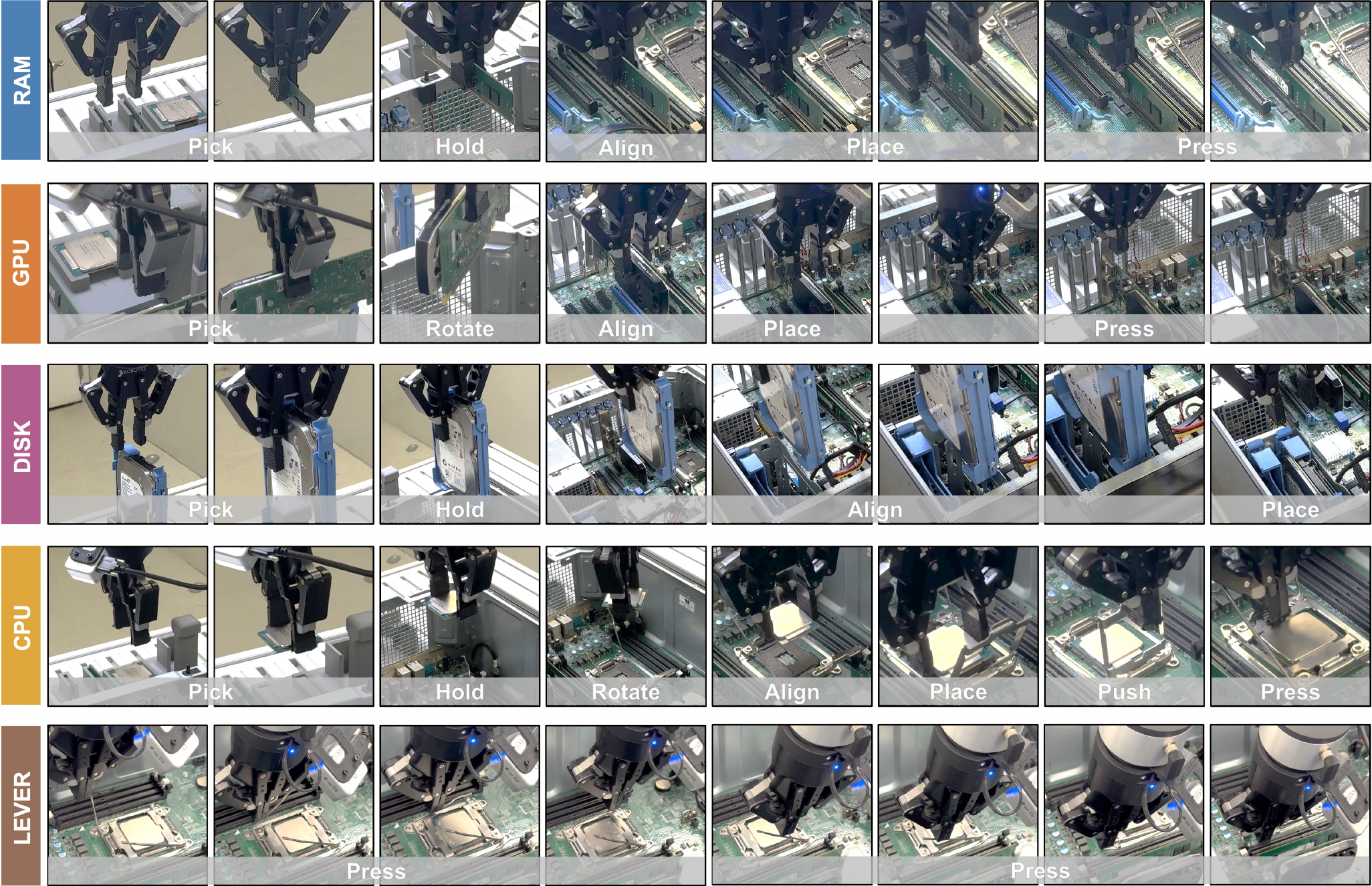}
  \caption{The five tasks of the assembly suite, each shown as the sequence of sub-goals the policy executes. Verbs are shared across tasks: \emph{pick}, \emph{hold} and \emph{rotate} are free-space, while \emph{align}, \emph{place}, \emph{push} and \emph{press} are contact-critical and hand control to the refinement expert. The LEVER task closes the CPU socket lever in two sequential presses and is the only task with no free-space sub-goal. Clearances and force limits are in Appendix~\ref{app:task_suite}, Table~\ref{tab:tasks}.}
  \label{fig:tasks}
\end{figure}

\textbf{Suite and protocol.} Five host-computer installation tasks share one chassis fixture and decompose into 23 sub-goals over a seven-verb vocabulary (Fig.~\ref{fig:tasks}): nine free-space, and fourteen contact-critical at 0.10--0.30\,mm clearance under per-phase safety limits $F_{\max}$. Chassis pose is randomized per trial, parts arrive by automated feeder, and every cell in the main tables is 20 trials under a pre-declared protocol with no re-runs.

\textbf{Metrics and baselines.} We report success, violation, intervention and recovery rates, peak off-axis force, cycle time and inference latency. The baselines are $\pi_{0.5}$ \cite{intelligence2025pi05}, $\pi_{0.5}$+F (wrench as an extra input token), $\pi_{0.5}$+RECAP-style (our implementation of the RECAP advantage-conditioning objective \cite{pi06recap2025} on the $\pi_{0.5}$ backbone), GR00T N1.7 \cite{nvidia2026gr00tn17} and TA-VLA \cite{tavla2025}, all trained on \dataset{} with matched budgets under one aligned protocol. For controlled \system{} variants, \emph{Align} learns the semantic--contact representation, \emph{+RL} adds value-guided post-training, and \emph{Full} additionally enables local adaptation. Appendix~\ref{app:protocol} gives task chains, clearances, metric definitions and baseline configurations.

\subsection{Main Comparison and Component Gains (Q1)}
\label{sec:exp_main}

Table~\ref{tab:task_success} reports task-level success for all eight methods, Fig.~\ref{fig:main_results} resolves four align-containing tasks over the reported diagnostic sub-goals, and Table~\ref{tab:main} in Appendix~\ref{app:results} gives the corresponding 17-row subset (13 contact-critical) of the 23-sub-goal suite. Averaged over the five tasks, \system{} attains 82\% success against 15\% for the strongest baseline, $\pi_{0.5}$+RECAP-style, a $5.5\times$ improvement. Per-task margins over the best baseline range from $+45$ points on the LEVER task, the only task without a free-space sub-goal, to $+75$ points on GPU, with the remaining three tasks between $+60$ and $+65$.

\begingroup
\hypersetup{citecolor=black}
\begin{figure}[t]
\centering
\captionsetup{hypcap=false,labelfont={bf,color=black},textfont={color=black}}
\captionof{table}{Task-level success rate (\%) on the assembly suite, 20 trials per cell. \textbf{Left:} all eight methods, with baseline references as given in Sec.~\ref{sec:exp_setup}; best results are bold and second-best results are underlined, with ties retained. The mean covers all five tasks. \textbf{Right:} representative alignment views for the four insertion tasks.}
\label{tab:task_success}
\color{black}
\begin{minipage}[t]{0.652\linewidth}
\vspace{0pt}
\raggedright
\setlength{\tabcolsep}{5pt}
\renewcommand{\arraystretch}{1.15}
\scriptsize
\begin{tabular}{@{}l ccccc @{\hspace{16pt}} c@{\hspace{16pt}}c@{\hspace{16pt}}c @{}}
\toprule
& \multicolumn{5}{c}{Baselines} & \multicolumn{3}{c@{}}{\system{} (ours)} \\
\cmidrule(lr){2-6}\cmidrule(l){7-9}
Task & $\pi_{0.5}$ & +F & +RECAP & GR00T & TA-VLA & Align & +RL & Full \\
\midrule
RAM  & 10 & 15 & 35 & 10 & 10 & 15 & \underline{45} & \textbf{95} \\
CPU  & 5  & 5  & 15 & 5  & 20 & 20 & \underline{45} & \textbf{85} \\
Disk & 25 & 20 & 20 & 5  & 30 & 30 & \underline{65} & \textbf{95} \\
GPU  & 10 & 5  & 5  & 0  & 5  & 10 & \underline{35} & \textbf{85} \\
LEVER & 0  & 0  & 0  & 0  & \underline{5} & \underline{5} & 0 & \textbf{50} \\
\midrule
Mean (5 tasks) & 10 & 9 & 15 & 4 & 14 & 16 & \underline{38} & \textbf{82} \\
\bottomrule
\end{tabular}
\end{minipage}\hfill
\begin{minipage}[t]{0.323\linewidth}
\vspace{0pt}
\centering
\includegraphics[width=\linewidth]{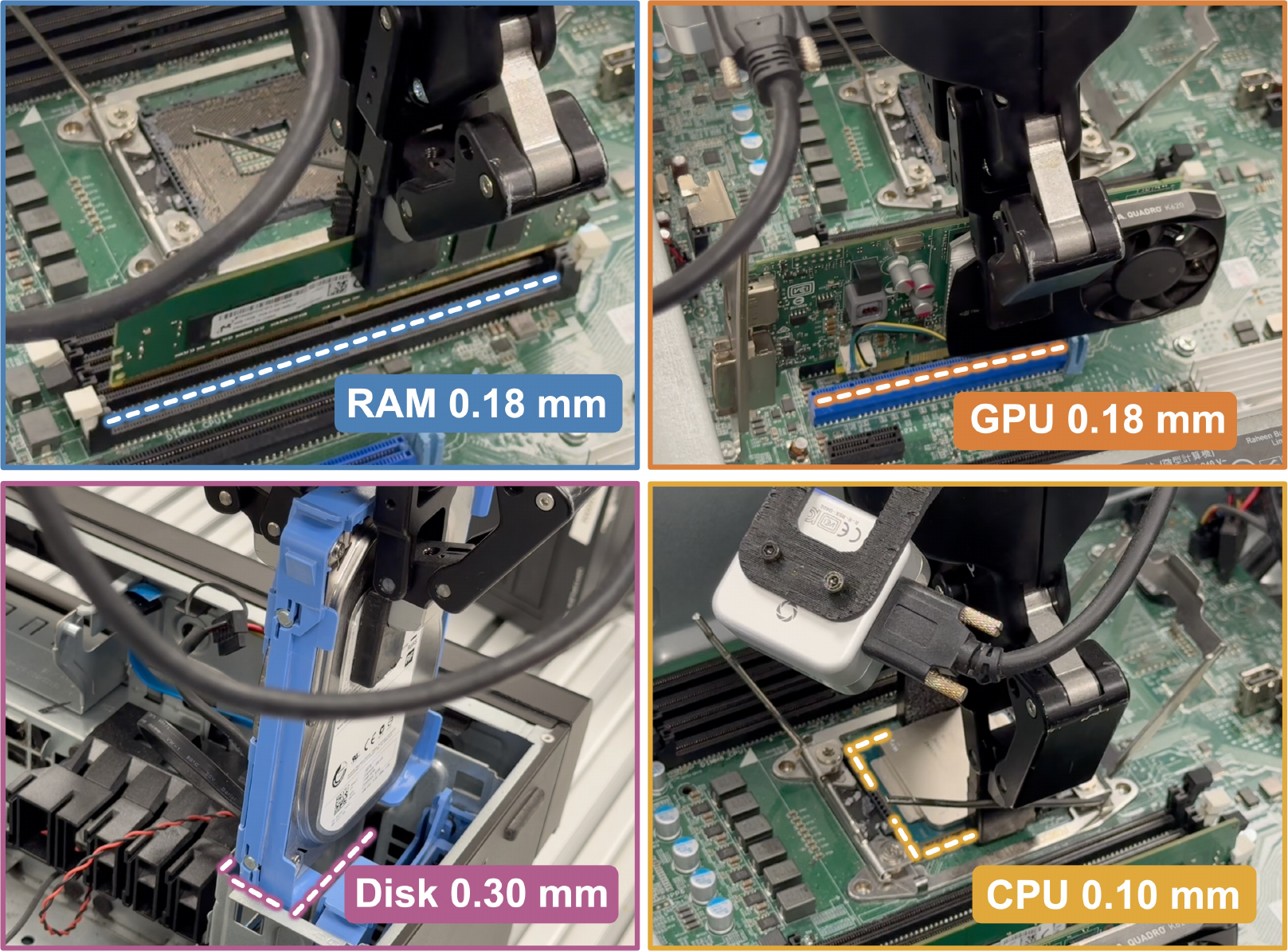}
\end{minipage}
\par\vspace{0.75em}
\includegraphics[width=0.97\linewidth]{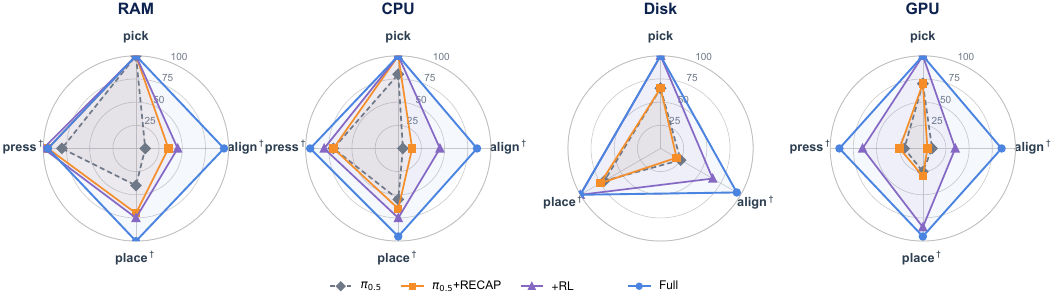}
\captionof{figure}{Sub-goal success on four insertion tasks. Performance remains high in free-space \emph{pick} and contracts at contact-critical sub-goals ($\dagger$); the two-step LEVER task is reported in Tables~\ref{tab:task_success} and~\ref{tab:main}.}
\label{fig:main_results}
\vspace{-0.65em}
\end{figure}
\endgroup

\textbf{Failures concentrate at contact.} No baseline exceeds 35\% on any task, and the diagnostic rows of Table~\ref{tab:main} locate the measured deficit: the displayed \emph{pick} rows run from 60\% to 100\% for every method, while \emph{align} and \emph{place} are where performance falls. Fig.~\ref{fig:main_results} shows the same shape task by task, with every method near the rim on \emph{pick} and the polygons moving inward on the sub-goals marked $\dagger$. RAM and GPU \emph{align} are both 10\% for $\pi_{0.5}$ against 95\% and 85\% for \system{}, localizing the largest gap to contact. Because a task succeeds only if every sub-goal in its chain succeeds, the task rows sit at or below the weakest link; the LEVER task, two contact-critical presses with no free-space step, is the hardest task in the suite and the only one where \system{} stays at 50\%.

\textbf{Feeling contact is not acting on it.} $\pi_{0.5}$+F and TA-VLA are the controls that matter for our central claim: both receive contact feedback, yet neither uses deployment value to distinguish the consequences of proposed actions. Both land in the 9--16\% band shared by $\pi_{0.5}$ and the \emph{Align} variant, five to nine times below the full system.

\textbf{Representation, value, and adaptation are complementary.} The controlled variants trace 16\% $\to$ 38\% $\to$ 82\% without changing the evaluation protocol. \emph{Align} establishes the joint action--wrench representation but has not yet used deployment outcomes to rank interactions. Adding value-guided RL contributes 22 mean points; its gains concentrate on the four insertion tasks, improving them by 25--35 points, while the contact-only LEVER task remains unresolved. Local adaptation supplies the remaining part-specific correction: relative to \emph{+RL}, the full model gains 50 points on RAM, 40 on CPU, 30 on Disk, 50 on GPU, and 50 on LEVER. Because task success requires the entire sub-goal chain, these gains measure completed assemblies rather than isolated contact events.

\textbf{The improvement is consistent across task geometry.} The full model reaches 85--95\% on the four insertion tasks despite different parts, approach motions, and 0.10--0.30\,mm clearances. LEVER remains the hardest at 50\% because both commands are contact-critical presses and either failure invalidates the trial. This pattern separates breadth from precision: generalist baselines handle the shared free-space vocabulary, while the aligned representation, deployment value, and local contact correction determine whether that motion completes an assembly. Table~\ref{tab:efficiency} in Appendix~\ref{app:supporting_results} adds the deployment properties: a command every 50\,ms against 150\,ms for $\pi_{0.5}$, enabling the 20\,Hz refinement expert, and a task completed 40\% faster than an expert teleoperator.

\textbf{Chain completion amplifies the contact bottleneck.} The full-system task rate matches the weakest displayed contact-critical row in every chain: 95\% for RAM and Disk, 85\% for CPU and GPU, and 50\% for LEVER (Table~\ref{tab:main}). By contrast, every displayed free-space \emph{pick} row is 100\%. Across the 13 reported contact sub-goals, the unweighted mean is 87\%, five points above the 82\% end-to-end mean. This gap follows from the chain criterion: every transition must succeed. It also explains why a bounded local correction at one limiting contact state can produce a much larger gain in completed assemblies.

\begin{takeaway}
Generalist VLAs retain strong performance on the displayed free-space \emph{pick} sub-goals, while the measured deficit concentrates at contact. Supplying the wrench as an input does not close it, since $\pi_{0.5}$+F and TA-VLA sit with $\pi_{0.5}$ and \emph{Align}. The semantic--contact representation, value-guided refinement, and local adaptation jointly address that deficit.
\end{takeaway}

\FloatBarrier

\subsection{Post-Training from Deployment Experience (Q2)}
\label{sec:exp_posttrain}

\begin{wrapfigure}[14]{r}{0.455\textwidth}
  \vspace{-0.7\baselineskip}
  \centering
  \includegraphics[width=\linewidth]{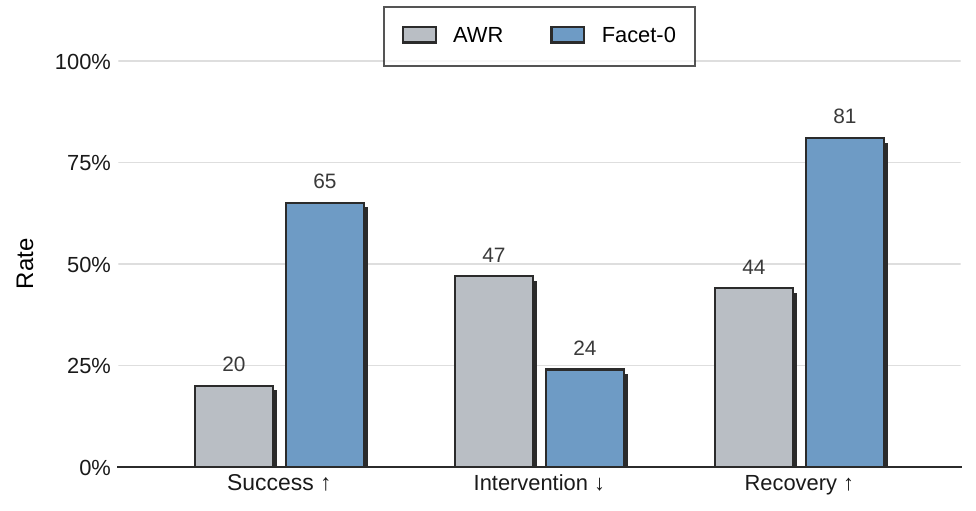}
  \captionsetup{hypcap=false,skip=5pt}
  \caption{Disk post-training with matched data and budget. Contact-selective RL raises success and recovery while reducing intervention. Arrows indicate preferred directions; Table~\ref{tab:posttrain} reports values.}
  \label{fig:posttrain_bar}
  \vspace{-0.4\baselineskip}
\end{wrapfigure}

Against matched advantage-weighted behavior cloning (AWR~\cite{peng2019advantage}) on the Disk deployment corpus, contact-selective post-training raises success from 20\% to 65\% and recovery from 44\% to 81\%. Human intervention falls from 47\% to 24\% (Fig.~\ref{fig:posttrain_bar}).

Unlike AWR, which reweights the deployment corpus by a scalar advantage, our contact-selective credit ranks samples within interaction regimes. It therefore retains high-credit contact-recovery frames that are scarce in demonstrations (Sec.~\ref{sec:value_refinement}; Appendix~\ref{app:behavior}).

\textbf{Joint operational effect.} Relative to AWR, contact-selective post-training adds 45 success points and 37 recovery points while reducing intervention by 23 points. The gain is therefore not purchased with additional human takeovers: improved completion coincides with lower supervision demand and a greater ability to continue after failure. This agreement across task outcome, operator involvement, and recovery provides stronger evidence of deployment reliability than any metric considered in isolation.

\begin{takeaway}
Critic-guided post-training acts on the tail of the deployment distribution rather than on the average trajectory. Contact failures that vision cannot resolve remain separable in the wrench, so recovery frames retain positive credit instead of being filtered out (Appendix~\ref{app:behavior}).
\end{takeaway}

\FloatBarrier

\subsection{Adaptation: Paradigms and Few-Shot Transfer (Q3)}
\label{sec:exp_adapt}

Adaptation is evaluated along two axes. The first asks whether the adaptation actor is the right \emph{mechanism} for on-robot learning, holding the task fixed. The second asks how cheaply the adapted system reaches an \emph{unseen part}.

\noindent
\begin{minipage}[t]{0.485\linewidth}
\vspace{0pt}
\textbf{Adaptation dynamics.} We first ask how the adaptation actor behaves as on-robot experience accumulates, against latent-noise steering of the frozen policy (DSRL \cite{wagenmaker2025dsrl}) under identical VLM reward. Fig.~\ref{fig:adaptation_ram_cpu} tracks the two quantities that decide whether on-robot learning is affordable, over adaptation episodes. \emph{Ours (TD3+BC)} is the adaptation actor of Sec.~\ref{sec:local_policy_adaptation}; \emph{Ours (HG-DAgger BC)} replaces its actor--critic objective with interactive imitation of the same interventions, holding the interface and the bounded parameterization fixed. DSRL stays at the floor on both tasks for the episode budget shown. Both \system{} variants reach the ceiling, and the actor--critic variant ends at full success on either task, but they arrive differently: the imitation variant starts high and is comparatively flat, whereas the actor--critic variant starts lower, passes through a pronounced dip on RAM insert, and recovers. Autonomy rises in the trailing mean for both, which is the property that makes the procedure affordable, since the supervisor is consulted less as the episode count grows.
\end{minipage}\hfill
\begin{minipage}[t]{0.48\linewidth}
  \vspace{0pt}
  \centering
  \includegraphics[width=\linewidth]{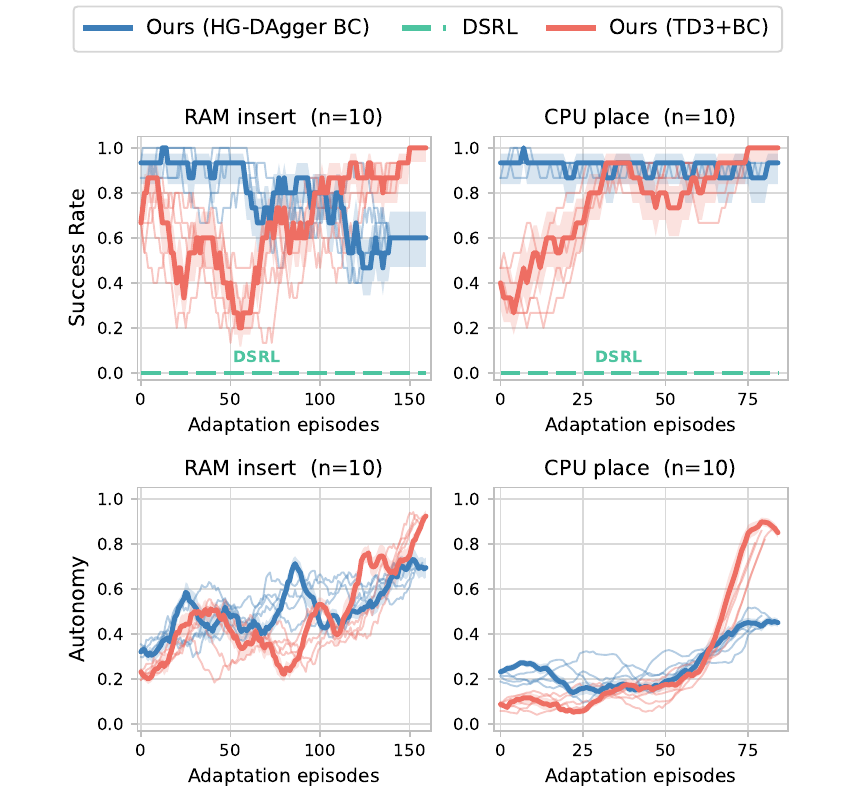}
  \captionsetup{hypcap=false}
  \captionof{figure}{On-robot adaptation dynamics, $n=10$ runs per curve. \textbf{Top}: success rate. \textbf{Bottom}: autonomy, the fraction of steps run without a supervisor takeover. Pale traces show representative individual runs; bold traces show a 15-episode trailing mean with a local $1\sigma$ band.}
  \label{fig:adaptation_ram_cpu}
\end{minipage}
\par\medskip

\begin{figure}[t]
\centering
\begin{minipage}[c]{0.370\linewidth}
  \centering
  \panelcard{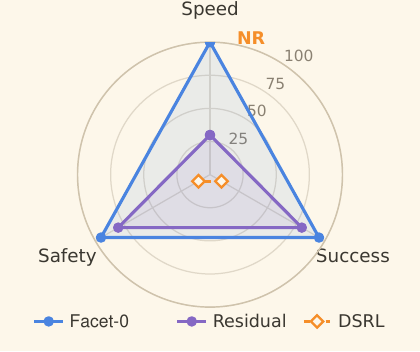}
\end{minipage}\hfill
\begin{minipage}[c]{0.605\linewidth}
  \centering
  \panelcard{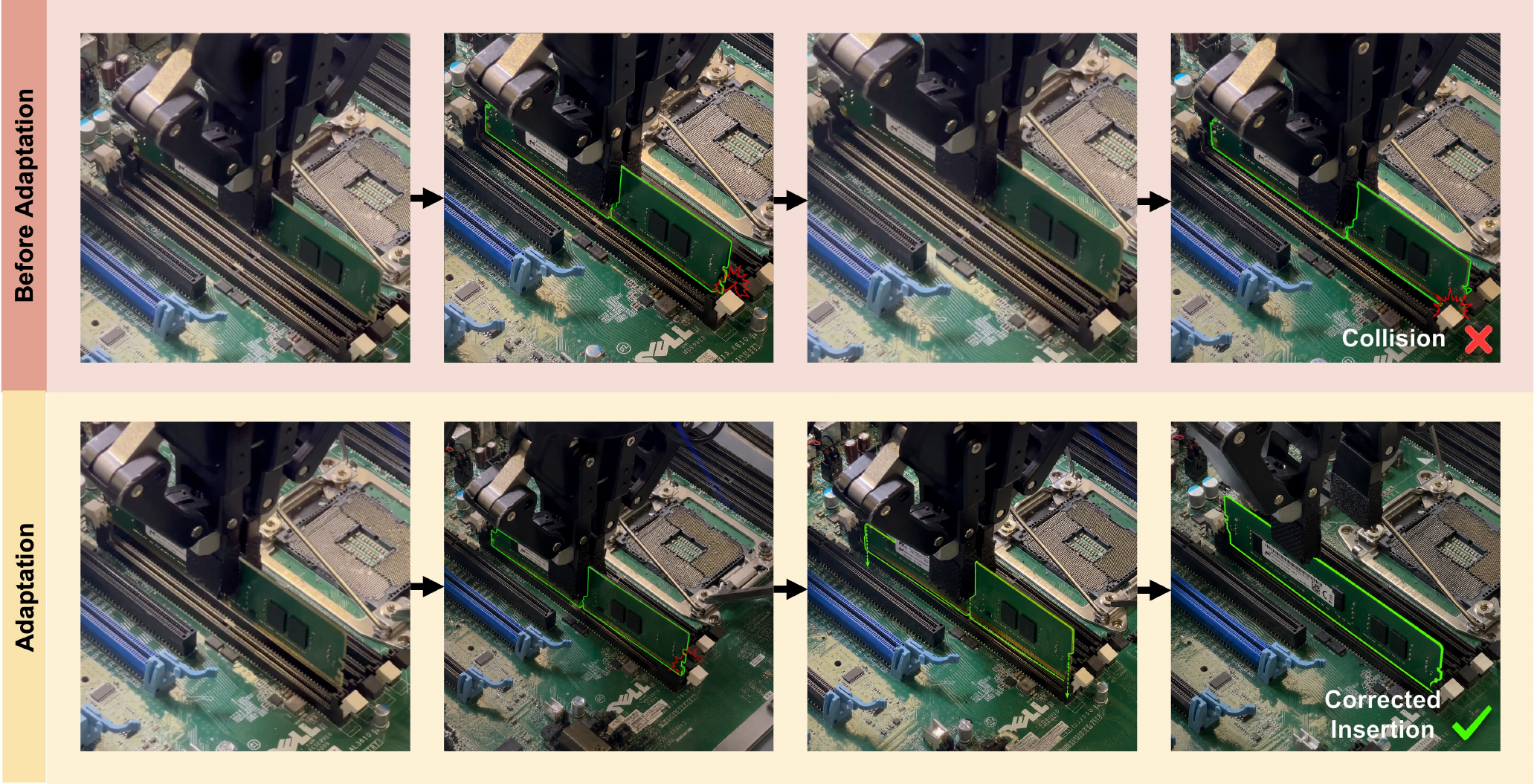}
\end{minipage}
\caption{Adaptation on RAM insertion. \textbf{Left}: normalized comparison with position-only residual adaptation and DSRL; speed is time to 90\% success, and NR means not reached. \textbf{Right}: after an off-axis collision, the adapted policy backs off, re-aligns and seats the module. Other axes and raw values are in Table~\ref{tab:adapt_paradigm}.}
\label{fig:adaptation}
\end{figure}

\textbf{Adaptation paradigms.} Fig.~\ref{fig:adaptation} summarizes this comparison at convergence and adds position-only residual adaptation \cite{residualoffpolicy2025}; raw measurements are in Appendix~\ref{app:supporting_results}, Table~\ref{tab:adapt_paradigm}. Position-only residuals correct alignment, whereas our bounded actor adds an auxiliary wrench-consequence objective that preserves action--contact coupling. This prediction is not commanded; the right panel shows the resulting behavior.

\textbf{Few-shot transfer to an unseen part.} Table~\ref{tab:adapt_fewshot} reports transfer to a memory module the system has never seen, with a different mass, stiffness, and latch signature; all methods receive ten demonstrations and three hours of training. \system{} reaches $9\times$ the strongest baseline while updating a fifteenth of the parameters, and the two facts are connected. The baselines must relearn the task from ten demonstrations because they have no interface that separates what changed, the contact dynamics of a heavier module with a stiffer latch, from what did not, the slot geometry, the approach, and the semantics of the instruction. \system{} trains only the adaptation actor, so the ten demonstrations are spent entirely on the part of the policy the unseen module actually invalidates.

\begin{takeaway}
Bounded action adaptation with auxiliary wrench prediction converts contact from a terminal failure into an actionable correction: after an off-axis collision the adapted policy backs off, re-aligns and seats the module on a second attempt (Fig.~\ref{fig:adaptation}). The executable correction comes from the action head; the wrench head remains predictive and non-commanded, and the refinement expert stays frozen.
\end{takeaway}

\FloatBarrier

\begin{table}[H]
\centering
\begin{minipage}[t]{0.425\linewidth}
\centering
\begin{minipage}[t][5\baselineskip][t]{\linewidth}
\caption{Few-shot transfer to an unseen RAM module with a different mass, stiffness and latch signature. All methods receive 10 demonstrations and three hours of training. \emph{Params.} is the fraction of policy parameters updated.}
\label{tab:adapt_fewshot}
\end{minipage}
\setlength{\tabcolsep}{5pt}
\renewcommand{\arraystretch}{1.12}
\footnotesize
\begin{tabular*}{\linewidth}{@{\extracolsep{\fill}}lcc@{}}
\toprule
Method & Success (\%) $\uparrow$ & Params. $\downarrow$ \\
\midrule
$\pi_{0.5}$ & 5 & 100\% \\
$\pi_{0.5}$+F & 5 & 100\% \\
GR00T N1.7 & 0 & 100\% \\
TA-VLA & 5 & 100\% \\
\rowcolor{gray!10}
\textbf{\system{}} & \textbf{45} & \textbf{6.6\%} \\
\bottomrule
\end{tabular*}
\end{minipage}\hfill
\begin{minipage}[t]{0.545\linewidth}
\centering
\begin{minipage}[t][5\baselineskip][t]{\linewidth}
\caption{Contact-role ablation on RAM insertion: the wrench signal is removed from prediction, interaction valuation, or local adaptation. The full model reproduces the RAM row of Table~\ref{tab:task_success}. Peak off-axis force is normalized to the force-free variant, so $1.0\times$ is the reference and lower is better.}
\label{tab:lifecycle}
\end{minipage}
\setlength{\tabcolsep}{5pt}
\renewcommand{\arraystretch}{1.12}
\footnotesize
\begin{tabular*}{\linewidth}{@{\extracolsep{\fill}}lcc@{}}
\toprule
Variant & Success (\%) $\uparrow$ & Peak force $\downarrow$ \\
\midrule
\rowcolor{gray!10}
\textbf{Full recipe} & \textbf{95} & $\mathbf{0.2\times}$ \\
w/o predictive wrench & 90 & $0.2\times$ \\
w/o critic wrench & 85 & $0.5\times$ \\
w/o adaptation wrench & 85 & $0.4\times$ \\
w/o all contact roles & 45 & $1.0\times$ \\
\bottomrule
\end{tabular*}
\end{minipage}
\end{table}

\subsection{Ablation Studies (Q4)}
\label{sec:exp_ablation}

\textbf{Contact-role ablation.} Table~\ref{tab:lifecycle} removes one action--wrench role at a time on RAM insertion. Joint removal is decisive: removing prediction, valuation, and adaptation cuts success by 50 points and returns peak off-axis force from $0.2\times$ to the force-free $1.0\times$. Individual removals differ by only one to two successes in 20 trials, so their ordering is not resolved, but the contact-quality pattern is consistent: predictive-wrench removal preserves the $0.2\times$ peak, whereas removing critic or adaptation wrench raises it to $0.5\times$ and $0.4\times$. Sec.~\ref{sec:discussion} states the scope of this comparison.

\begin{takeaway}
Removing every predictive and evaluative role of wrench is the one degradation that is large relative to the measurement resolution. Individual removals trend consistently with complementary functions, but at 20 trials per cell they span only one to two trials and do not establish a ranking among prediction, valuation, and adaptation.
\end{takeaway}

\FloatBarrier

\section{Discussion and Limitations}
\label{sec:discussion}

The training corpus spans three embodiments and two chassis families, while the controlled five-task evaluation uses wrist-sensed, parallel-gripper electronics assembly on a shared chassis fixture. Within this setting, the results support semantic--contact representation learning, value-guided refinement, and bounded adaptation, and the contact-role ablation distinguishes the complementary roles of prediction, valuation, and adaptation. Future work will broaden controlled evaluation to further end effectors, parts, and contact geometries, and explore force-conditioned policy distillation for cells without wrist-mounted six-axis sensing \cite{fdvla2026}.

\section{Conclusion}
\label{sec:conclusion}

This work presents \system{}, a robotic foundation model that links task semantics with anticipated contact and uses value-guided RL post-training to refine the behavior expressed through that representation. Its force-synchronized data, joint action--wrench prediction, Action--Wrench Critic, and bounded local adaptation connect perception, credit assignment, and execution at contact. On the five-task electronics-assembly suite, the full model reaches 82\% mean success (15\% for the strongest baseline), with 0.5\,mm placement accuracy and 50\,ms command latency. Controlled variants obtain 16\% with semantic--contact alignment, 38\% after value-guided refinement, and 82\% with local adaptation. Relative to the matched AWR baseline, value-guided post-training reduces intervention from 47\% to 24\% and increases recovery from 44\% to 81\%.

The results indicate that one persistent semantic--contact representation can support both global policy improvement and local transfer: the Action--Wrench Critic values the policy's joint proposal for global refinement, while a downstream bounded actor reuses the frozen representation and policy reference for an unseen module with ten demonstrations while updating 6.6\% of the parameters. These findings are established on wrist-sensed electronics assembly, and extending them to further embodiments and contact regimes is ongoing work. We release the model, dataset, and training recipe to support that evaluation.

\begingroup
\small
\bibliographystyle{plain}
\bibliography{references}
\endgroup

\newpage

\appendix
\section{Assembly Suite Details}
\label{app:task_suite}

The suite uses a shared chassis fixture but exposes distinct contact geometries. Fig.~\ref{fig:tasks} in the body records each execution chain; Table~\ref{tab:tasks} gives the evaluated clearances and configured force limits. LEVER is evaluated as an isolated, contact-only CPU socket lever closure rather than as another full pick-to-place chain.

\begin{table}[H]
\centering
\caption{The host-computer assembly suite: 23 sub-tasks, fourteen of them contact-critical. Sub-goals are listed in execution order. Clearance is the tightest fit among the task's critical sub-goals. $F_{\max}$ is the configured safety limit for the indicated contact phase; CPU uses 8\,N for pin-array seating and 35\,N for socket closure.}
\label{tab:tasks}
\setlength{\tabcolsep}{3.5pt}
\renewcommand{\arraystretch}{1.15}
\scriptsize
\begin{tabular}{@{}llccl@{}}
\toprule
Task & Sub-goal sequence (free-space, \textbf{contact-critical}) & Clear.\ (mm) & $F_{\max}$ (N) & Contact character \\
\midrule
RAM install & pick, hold, \textbf{align}, \textbf{place}, \textbf{press} & 0.18 & 70 & sequential per-end latch presses \\
CPU install & pick, hold, rotate, \textbf{align}, \textbf{place}, \textbf{push}, \textbf{press} & 0.10 & 8 / 35 & fragile pin-array seating; ZIF closure \\
Disk install & pick, hold, \textbf{align}, \textbf{place} & 0.30 & 20 & guided bay seating \\
GPU install & pick, rotate, \textbf{align}, \textbf{place}, \textbf{press} & 0.18 & 55 & long-edge PCIe alignment; latch click \\
LEVER closure & \textbf{press}, \textbf{press} & 0.10 & 60 & two sequential lever presses \\
\bottomrule
\end{tabular}
\end{table}

\subsection{Evaluation Protocol}
\label{app:protocol}

This subsection restores the setup detail compressed out of Sec.~\ref{sec:exp_setup}.

\textbf{Task suite.} Each task carries an overall instruction, for example \emph{pick up the RAM and install it}, which the backbone decomposes into the sub-goals shown in Fig.~\ref{fig:tasks}, for example \emph{press down the RAM until it locks into place}. The five tasks comprise 23 sub-tasks drawn from a shared seven-verb vocabulary (\emph{pick}, \emph{hold}, \emph{rotate}, \emph{align}, \emph{place}, \emph{push}, \emph{press}), so a skill learned in one task is addressable in another. Nine sub-tasks are free-space (\emph{pick}, \emph{hold}, \emph{rotate}) and execute on the coarse chunk alone; the remaining fourteen are contact-critical (\emph{align}, \emph{place}, \emph{push}, \emph{press}) and activate the refinement expert. Clearances range from 0.10 to 0.30\,mm, and each contact phase uses a configured safety limit $F_{\max}$; Fig.~\ref{fig:tasks} and Table~\ref{tab:tasks} give the full sequences, clearances and limits.

\textbf{Protocol.} Chassis pose is randomized per trial and parts are supplied by an automated feeder. Each cell in the main tables is 20 trials under a protocol declared before evaluation, with no re-runs.

\textbf{Metrics.} \emph{Success rate}: the installation completed without force violation. \emph{Peak off-axis force}: the maximum contact force orthogonal to the task-frame motion axis, which measures misalignment damage risk rather than deliberate seating force. \emph{Violation rate}: the fraction of trials with any control step above $F_{\max}$. \emph{Intervention rate}: the fraction of trials in which a human supervisor took over. \emph{Recovery rate}: the fraction of detected failures the system resolved on its own. \emph{Cycle time}: wall-clock per successful trial. \emph{Inference latency}: wall-clock from observation to command.

\textbf{Baselines.} All baselines are trained on \dataset{} with matched budgets and evaluated under one aligned protocol. $\pi_{0.5}$ \cite{intelligence2025pi05} uses the same vision--language observations and 7-D kinematic state but no wrench channel. $\pi_{0.5}$+F adds the wrench as an input token and changes nothing else, which isolates force as an input from force as a lifecycle commitment. $\pi_{0.5}$+RECAP-style applies the published RECAP value-estimation and advantage-conditioning objective \cite{pi06recap2025} to the same $\pi_{0.5}$ backbone; it is our aligned reimplementation rather than the released $\pi_{0.6}^{\star}$ model. GR00T N1.7 \cite{nvidia2026gr00tn17} is a second generalist backbone of comparable scale. TA-VLA \cite{tavla2025} is an aligned reimplementation on the same backbone that aggregates wrist F/T history into a single token. The \emph{Align} and \emph{+RL} variants of \system{} denote semantic--contact alignment alone and its value-guided refinement, respectively; \emph{Full} additionally includes local policy adaptation.

\FloatBarrier

\section{Dataset Curation Protocol}
\label{app:dataset_curation}

This section expands the collection and curation summary of Sec.~\ref{sec:dataset}. The acquisition network uses a shared sensing specification: each cell has a torque-controlled arm, parallel gripper, wrist six-axis force/torque sensor, three RGB cameras, and joint proprioception. Kinesthetic teleoperation with force feedback is used for demonstrations. Hardware timestamps on a shared clock define a common 15\,Hz training timeline; high-rate state and wrench streams are retained for the 200\,Hz controller. Episodes whose cross-channel skew exceeds tolerance are rejected rather than interpolated.

\textbf{Validity extraction.} Teleoperation faults, truncated chunks, sensor saturation, calibration drift beyond bounds, and sustained manipulated-part occlusion are removed at segment level, preserving unaffected intervals of an episode. Action chunks are scored by action and wrench variation over the horizon; near-stationary pauses, repositioning and fixture waits are down-sampled to avoid duration-dominated mixtures.

\textbf{Semantic and phase annotation.} Each segment receives two VLM-generated question--answer pairs: its overall task/current sub-task and the next instruction (e.g., ``press down the RAM until it locks into place''). Annotators mark boundaries for the seven phases (\textit{approach}, \textit{align}, \textit{insert}, \textit{press}, \textit{seat}, \textit{fasten}, \textit{retreat}) using keyboard chords; the VLM pass verifies boundaries and flags disagreements for re-annotation. The resulting labels are reused verbatim by pretraining and reward decomposition.

\textbf{Anomaly reflow.} Trajectories with anomalous jumps are paired with the corrected trajectory that follows and both are retained. This preserves failure and recovery signal for post-training while allowing the curation log to trace each correction. Center identities are pseudonymised, personally identifying content is removed, and the release includes the data card and curation code.

\FloatBarrier

\section{Architecture and Execution Details}
\label{app:arch}

This appendix records the engineering realization of the architecture of Sec.~\ref{sec:arch}: the execution hierarchy, the compliant controller, the safety operator, and the design rationale compressed out of the main text. Nothing here changes the model definition.

\textbf{Why the wrench history is a window.} A single wrench sample is ambiguous. Contact onset is a \emph{step} in the wrench, a jam is a wrench rising while displacement stalls, and a skew is an off-axis moment persisting under axial push; each is a property of a short trajectory, not of one frame, and none is visible to vision once the part occludes the slot. $K=10$ frames is the shortest window over which the contact-onset step is detectable above sensor noise at the corpus sampling rate.

\begin{figure}[H]
  \centering
  \includegraphics[width=\linewidth]{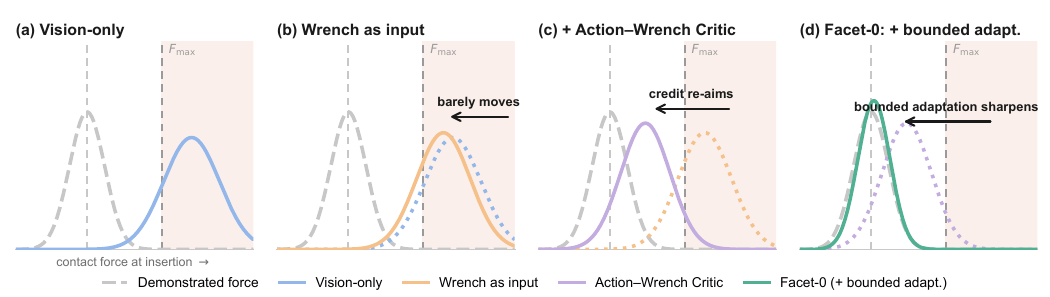}
  \caption{Schematic of how the action--wrench representation is intended to shape contact behavior across optimization. The distributions illustrate the roles of input conditioning, Action--Wrench valuation, and bounded local adaptation; they are conceptual rather than empirical measurements.}
  \label{fig:force_roles}
\end{figure}

\textbf{Why the wrench is predicted rather than only read.} Conditioning an action on the current wrench hands the policy a record of contact its own past actions already produced: it arrives too late to shape the action that caused it, and nothing in a behavior-cloning loss rewards attending to six force dimensions among thousands of visual ones. Making the wrench a decoding target changes the learning signal: the wrench term in Eq.~\eqref{eq:awloss} falls only for a model that represents the contact consequence of its proposed motion, and $\widehat{W}$ becomes a predictive variable available before execution. Fig.~\ref{fig:force_roles} schematically summarizes the intended roles of this variable across training.

\textbf{Execution hierarchy.} The joint action--wrench proposal of Sec.~\ref{sec:arch} is decoded at $5$--$10$\,Hz at roughly $5$\,mm resolution. A contact-rate refinement expert consumes that chunk together with $h_t^c$ and returns a refined action--wrench chunk at $20$\,Hz at roughly $0.5$\,mm; coarse and fine differ in precision, not in kind. Beneath both, a compliant controller runs at $200$\,Hz. The refinement expert is frozen during the local policy adaptation of Sec.~\ref{sec:local_policy_adaptation}.

\textbf{Compliant controller.} The $200$\,Hz layer regulates force along the task-frame directions designated by the sub-goal and tracks position along the remainder \cite{raibert1981hybrid, mason1981compliance, hogan1985impedance, beltran2020admittance}. Its force loop closes on measured wrist F/T feedback. Predicted wrench is retained as a supervised contact-consequence variable for representation and value learning; it is neither a controller input nor a force setpoint.

\textbf{Safety operator.} $\mathcal{S}_{\mathrm{task}}$ sits between policy and controller and clips the commanded target to the workspace, limits per-step motion, canonicalizes or passes through rotation, optionally smooths in $SE(3)$, gates spatially, and passes the gripper command through unchanged. It is architectural rather than behavioral: a policy \emph{trained} to be gentle is gentle on its training distribution, whereas a clipped and rate-limited command is constrained on every distribution, including those that on-robot adaptation, mis-timed sub-goal boundaries, and unseen parts produce. Its guarantee is kinematic and we state its limit plainly: bounding displacement per step bounds how far a mis-aimed target can drive the tool between observations, which is what keeps early adaptation episodes cheap, but it is not the passivity bound of energy-tank formulations \cite{ferraguti2013tank, kronander2016stability}. The reward and compliant controller discourage excessive interaction force, but no certified hard force bound is claimed.

\textbf{Why adapt behind an interface rather than fine-tune the backbone.} Three reasons, in increasing order of importance. It is cheap: adaptation to an unseen part touches only the local actor and finishes in hours on one robot. It is safe in the regression sense: the backbone carrying open-world semantics is never updated by a few hundred contact episodes, so adding a skill cannot silently degrade the others. And it is safe in the operational sense: the actor's output is confined to the per-task admissible box $[a_{\min},a_{\max}]$ of Sec.~\ref{sec:local_policy_adaptation} and then passed through $\mathcal{S}_{\mathrm{task}}$, so every command is admissible by construction regardless of how far the actor departs from the frozen reference. We adopt a bounded absolute parameterization rather than a residual one precisely because a residual's safety comes from the smallness of its corrections, and that same smallness bounds departure from a base whose failures adaptation exists to repair.

\textbf{Predictive coupling across optimization.} Semantic--contact alignment and value-guided refinement operate on the joint action--wrench proposal $Y_t$ introduced in Sec.~\ref{sec:arch}. Local policy adaptation retains the same predictive relation through a bounded action head and a supervised one-step wrench head. Its twin critics and robot interface remain defined only over the executable Cartesian action; the wrench prediction is auxiliary and non-commanded.

\section{Method Details}
\label{app:method_details}

This appendix collects the optimization procedures, constants, and derivations deferred from Secs.~\ref{sec:method} and~\ref{sec:value_refinement}.

The three method modules share the semantic--contact representation $h_t^c$ and preserve its predictive action--contact relation. Semantic--contact alignment learns $h_t^c$ and decodes the joint proposal $Y_t=[A_{t:t+H-1},\widehat W_{t+1:t+H}]$ from demonstrations; value-guided policy refinement evaluates that proposal with the Action--Wrench Critic; and local policy adaptation compresses $h_t^c$ into $z_t$ while adapting only the executable action to a new part (Table~\ref{tab:method_modules}). Throughout, $\Omega$ denotes the VLM judge and $F_{\max}(\varphi)$ the force limit configured for contact phase $\varphi$.

\begin{table}[H]
\centering
\caption{The three method modules share $h_t^c$ while retaining predictive coupling between action and contact consequence. $Y_t$ denotes the decoded joint proposal, not the representation itself.}
\label{tab:method_modules}
\setlength{\tabcolsep}{3.2pt}
\renewcommand{\arraystretch}{1.2}
\scriptsize
\begin{tabular}{@{}>{\raggedright\arraybackslash}p{0.16\linewidth}
                    >{\raggedright\arraybackslash}p{0.21\linewidth}
                    >{\raggedright\arraybackslash}p{0.17\linewidth}
                    >{\raggedright\arraybackslash}p{0.18\linewidth}
                    >{\raggedright\arraybackslash}p{0.14\linewidth}@{}}
\toprule
Module & Representation / proposal role & Trainable & Frozen & Data \\
\midrule
Semantic--contact alignment & learns $h_t^c$; decodes $Y_t$ & $\pi_{\mathrm{align}}$ & -- & \dataset{} demos \\
Value-guided refinement & values $Y_t$; retains $h_t^c$ & $\pi_{\mathrm{ref}}$, $Q^{\mathrm{AW}}_\psi$, $E_z$ & VL backbone & rollouts + interventions \\
Local policy adaptation & compresses $h_t^c$; bounded heads & local actor $\eta$, $Q_\xi$ & $\pi_{\mathrm{ref}}$, $E_z$, $\Omega$ & on-robot experience \\
\bottomrule
\end{tabular}
\end{table}

\subsection{Semantic--Contact Alignment}
\label{app:semantic_contact_alignment}

\textbf{Masking and normalization.} $M^{a}$ and $M^{w}$ mask padded and invalid timesteps in the action and wrench slices of $Y_t$, respectively. Each term of Eq.~\eqref{eq:awloss} is normalized by its own mask sum rather than by a shared denominator, so a batch with few valid contact frames does not silently down-weight the force term.

\textbf{Optimizer alternation and update budget.} Algorithm~\ref{alg:semantic_contact_alignment} selects the action--wrench branch on four updates in five, giving 32K action-branch updates against 8K VQA updates on one run.

\textbf{Why the losses are never summed.} $\mathcal{L}_{\mathrm{AW}}$ is a flow-matching MSE and $\mathcal{L}_{\mathrm{VQA}}$ a token cross-entropy; they carry different units and differ by orders of magnitude, so a single weighted sum would make the effective weighting an artifact of scale rather than a design choice. Alternating two optimizers four-to-one keeps each branch's step size independent.

\textbf{Mixture pyramid.} The three tiers are fundamental operations (\emph{pick}, \emph{hold}, \emph{rotate}), precision operations (\emph{align}, \emph{place}, \emph{push}, \emph{press}), and long-tail recovery episodes. Sampling weights over-represent the upper two tiers relative to their raw duration, because duration and decisiveness are inversely related in assembly: free-space transit dominates wall-clock time and decides nothing.

\begin{algorithm}[H]
\caption{Semantic--contact alignment (Eq.~\eqref{eq:awloss})}
\label{alg:semantic_contact_alignment}
\begin{algorithmic}[1]
\Require demonstrations $\mathcal{D}_{\mathrm{pre}}$, VQA set $\mathcal{D}_{\mathrm{vqa}}$, pretrained VL backbone, mixture weights over \{base, precision, recovery\}
\Ensure aligned policy $\pi_{\mathrm{align}}$ generating $Y_t = [A_{t:t+H-1}, \widehat{W}_{t+1:t+H}] \in \mathbb{R}^{H \times 13}$
\For{update $k = 0, 1, 2, \dots$}
  \If{$k \bmod 5 \in \{0,1,2,3\}$} \Comment{action--wrench branch}
    \State sample $(I^{1:3}_t, \ell, s^{\mathrm{kin}}_t, W_{t-K+1:t}, Y_t^{\mathrm{data}})$ from $\mathcal{D}_{\mathrm{pre}}$ under the mixture pyramid
    \State $h^{\mathrm{VL}}_t\leftarrow E_{\mathrm{VL}}(I_t^{1:3},\ell)$;\quad $h^{c}_t \leftarrow F_{\mathrm{fuse}}(h^{\mathrm{VL}}_t,E_s(s^{\mathrm{kin}}_t),E_w(W_{t-K+1:t}))$ \Comment{Eq.~\eqref{eq:fusion}}
    \State $\epsilon \sim \mathcal{N}(0, I)$;\quad $\tau \sim \mathrm{Beta}(1.5, 1)$
    \State $X_\tau \leftarrow (1-\tau) Y_t^{\mathrm{data}} + \tau \epsilon$;\quad $U_\tau \leftarrow \epsilon - Y_t^{\mathrm{data}}$
    \State $v \leftarrow v_\theta(X_\tau, \tau \mid h^{c}_t)$
    \State $\mathcal{L}_{\mathrm{act}} \leftarrow \sum M^{a} \odot \lVert v^{a} - U_\tau^{a} \rVert_2^2 \,/\, \sum M^{a}$ \Comment{action columns}
    \State $\mathcal{L}_{\mathrm{wr}} \leftarrow \sum M^{w} \odot \lVert v^{w} - U_\tau^{w} \rVert_2^2 \,/\, \sum M^{w}$ \Comment{wrench columns}
    \State $\theta \leftarrow \mathrm{Opt}_{\mathrm{AW}}\big(\theta, \nabla \mathcal{L}_{\mathrm{AW}}\big)$, \ $\mathcal{L}_{\mathrm{AW}} = \mathcal{L}_{\mathrm{act}} + \lambda_{\mathrm{pre}} \mathcal{L}_{\mathrm{wr}}$ \Comment{$\lambda_{\mathrm{pre}} = 0.1$}
  \Else \Comment{VQA branch}
    \State sample $(I, q, y_{1:L})$ from $\mathcal{D}_{\mathrm{vqa}}$
    \State $\mathcal{L}_{\mathrm{VQA}} \leftarrow -\big(\textstyle\sum_i m_i\big)^{-1} \sum_i m_i \log p_{\theta_{\mathrm{VLM}}}(y_i \mid I, q, y_{<i})$
    \State $\theta \leftarrow \mathrm{Opt}_{\mathrm{VQA}}\big(\theta, \nabla \mathcal{L}_{\mathrm{VQA}}\big)$
  \EndIf
\EndFor
\end{algorithmic}
\end{algorithm}

\textbf{Alignment contract across the two branches.} Alternating optimization is useful only if the branches improve a common representation rather than exchange targets. The structured mask therefore exposes the same visual--language prefix to both branches while keeping their supervised outputs disjoint. The action--wrench branch additionally receives kinematic state, causal wrench history, and the noised joint chunk; the VQA branch receives preceding answer tokens. Neither branch can attend to the other's target tokens. Semantic supervision can consequently shape the shared prefix without revealing the answer sequence to the policy, while interaction supervision can attach physical consequence to that prefix without turning wrench into a language label.

\textbf{Temporal pairing and causality.} Row $k$ of $Y_t^{\mathrm{data}}$ pairs the command at $t+k$ with the measured wrench at $t+k+1$. This one-step offset is the smallest causal convention that assigns contact to the motion that produced it. Pairing an action with the contemporaneous pre-action sample would instead teach the decoder to reconstruct contact already present in the observation. Invalid or padded future samples are removed by $M^w$, so the offset changes neither the action horizon nor the number of executable commands. The resulting representation summarizes what the scene means, what the robot is doing, and what contact that motion is expected to produce.

\subsubsection{Held-out Alignment Diagnostics}
\label{app:alignment_diagnostics}

We diagnose alignment on held-out \dataset{} episodes before closed-loop evaluation. These plots compare checkpoint-level optimization under a matched update budget; they report validation losses rather than task success, and each task can attain its minimum at a different checkpoint.

\begin{figure}[H]
  \centering
  \includegraphics[width=\linewidth]{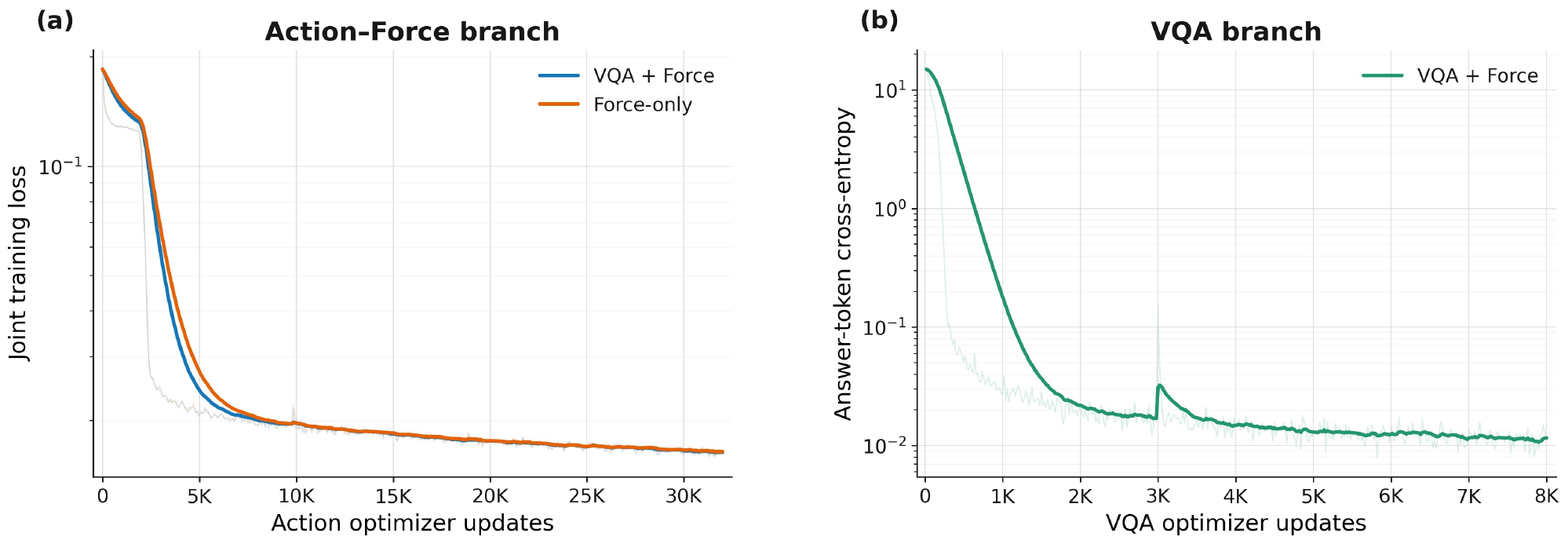}
  \caption{Training dynamics under a matched budget of 32K action-branch updates. \textbf{(a)} Joint action--wrench loss for VQA\,+\,Force and Force-only training. \textbf{(b)} Answer-token cross-entropy for the VQA branch over its 8K updates. Pale curves show raw values and solid curves show exponential moving averages.}
  \label{fig:pretrain_dynamics}
\end{figure}

\begin{figure}[H]
  \centering
  \includegraphics[width=\linewidth]{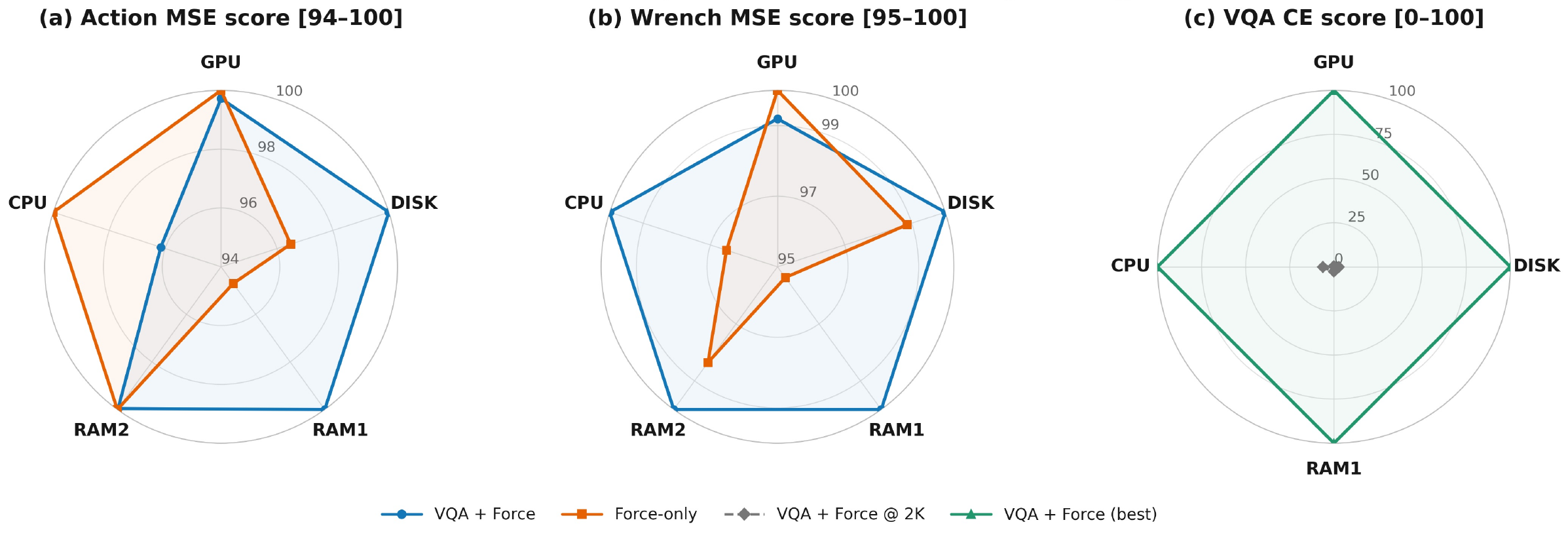}
  \caption{Best per-task validation loss within the matched budget, normalized as a higher-is-better score. \textbf{(a)} Action MSE and \textbf{(b)} wrench MSE use zoomed radial ranges of $[94,100]$ and $[95,100]$. \textbf{(c)} VQA cross-entropy uses the full $[0,100]$ range and compares the best joint checkpoint with the same model at 2K updates. RAM1 and RAM2 denote the two memory slots.}
  \label{fig:pretrain_radar}
\end{figure}

Figure~\ref{fig:pretrain_dynamics} compares the joint VQA-plus-wrench model with wrench-only training. Their action--wrench losses converge to nearly the same endpoint, while the answer-token cross-entropy of the joint model falls by roughly three orders of magnitude. Figure~\ref{fig:pretrain_radar} resolves the best validation losses by task. Its action and wrench axes are deliberately zoomed, so the visible motor differences are only a few percentage points, whereas the VQA comparison uses the full range. Together, the diagnostics show that semantic supervision is acquired without a meaningful degradation of action or wrench prediction.

\FloatBarrier

\subsection{Value-Guided Policy Refinement}
\label{app:value_guided_refinement}

\begin{algorithm}[H]
\caption{Value-guided policy refinement with the Action--Wrench Critic}
\label{alg:value_guided_refinement}
\begin{algorithmic}[1]
\Require aligned policy $\pi_{\mathrm{align}}$, task set $\mathcal{T}$, VLM judge $\Omega$
\Ensure value-refined policy $\pi_{\mathrm{ref}}$, Action--Wrench Critic $Q^{\mathrm{AW}}_\psi$, bottleneck encoder $E_z$
\State initialize $\pi_{\mathrm{ref}} \leftarrow \pi_{\mathrm{align}}$
\Repeat
    \State $\mathcal{D} \leftarrow \mathcal{D} \cup \mathrm{rollout}(\pi_{\mathrm{ref}}, \mathcal{T}) \cup \text{supervisor interventions}$
    \State $r_t \leftarrow$ Eq.~\eqref{eq:reward} under $\Omega$'s phase segmentation;\quad $c_t \leftarrow$ smoothed contact intensity
    \State train $Q^{\mathrm{AW}}_\psi$ by Eq.~\eqref{eq:critic}: distributional TD, progress-calibrated, four aux heads
    \State $\delta_t^{(N)} \leftarrow$ Eq.~\eqref{eq:advantage};\quad $b_t \leftarrow$ within-regime threshold \Comment{contact-enriched}
    \State retain $Y_t^{\mathrm{data}}=[A_{t:t+H-1}^{\mathrm{data}},W_{t+1:t+H}^{\mathrm{data}}]$ with contact-validity masking
    \State flow-matching update of $\pi_{\mathrm{ref}}$ by Eq.~\eqref{eq:rpt}: all frames train the unconditional branch, frames with $b_t=1$ additionally train the conditional branch.
\Until{round budget exhausted}
\State train bottleneck encoder $E_z$ on frozen backbone embeddings \Comment{used for local policy adaptation}
\end{algorithmic}
\end{algorithm}

\textbf{Distributional parameterization and auxiliary heads.} $Z_\psi$ of Eq.~\eqref{eq:critic} is a distributional
return model \cite{bellemare2017distributional} trained on all rollouts under progress-calibrated targets. The concrete
distributional parameterization and regression distance are implementation-dependent; the method requires only a return distribution whose mean defines $Q_\psi$. Four
auxiliary heads share the critic trunk and are trained jointly with the return objective: near-future wrench, contact intensity
$c_t$, within-contact progress, and success ranking. They exist to separate observations that agree on task progress but
disagree on contact state; without that separation the credit score of Eq.~\eqref{eq:advantage} cannot express a preference
between gentle alignment and ramming, because at equal progress the two are the same point.

\textbf{Wrench supervision during refinement.} Value-guided refinement does not introduce a new wrench variable. It retains the action-to-next-wrench pairing learned during semantic--contact alignment: each action row is supervised by the measured wrist wrench at the following step. Causal filtering, per-axis normalization, and a contact-validity mask are applied only to stabilize the loss; after de-normalization, $\widehat W_{t+1:t+H}$ retains the physical interpretation given in Sec.~\ref{sec:arch}. The executable action columns are unchanged.

\textbf{Contact regime and positive-label budget.} The regime $\rho_t$ of Eq.~\eqref{eq:rpt} is obtained by thresholding the contact intensity $c_t$, and
$q_{\rho}$ is the score threshold taken \emph{within} regime $\rho$ at a matched overall positive rate. Ranking within
regime rather than globally prevents the positive set from being dominated by free-space progress.

\textbf{Why the credit score is undiscounted and short.} Under a progress-calibrated critic, $V_\psi$ is approximately affine
in normalized task progress. A discounted bootstrap over the full chunk horizon therefore makes the score affine in
$V_\psi(h^c_t)$ itself, so early frames of every successful trajectory score highly merely because they are early, and the
positive-label budget collects on the free-space approach the policy already performs well. Truncating to $N<H$ and dropping
the discount makes $\delta_t^{(N)}$ measure local deviation from the trajectory's own calibrated progress rate, which is
invariant to where in the trajectory the frame sits.

\textbf{Prompt-level conditioning.} The positive-interaction label $b_t$ is appended to the instruction as a short literal tag,
so the conditional and unconditional branches of Eq.~\eqref{eq:rpt} differ only in that suffix and no architectural change is
required. At inference the two combine in classifier-free-guidance style \cite{ho2022cfg}; the guidance scale is a deployment
dial and $s=0$ recovers the unconditional policy exactly. Branch ratios are set so the unconditional branch always sees the
full data distribution and continues to act as a behavior-cloning anchor.

\textbf{Contact intensity.} $c_t \in [0,1]$ is computed from causally smoothed wrench differences rather than raw per-step deltas. Raw deltas in the high-rate sensor stream are dominated by noise; the smoothing window is the shortest that makes the contact-onset step detectable above sensor noise.

\subsection{Local Policy Adaptation}
\label{app:local_policy_adaptation}

\textbf{FACET bottleneck.} On task data, $E_z$ learns a frozen, per-valid-token reconstruction bottleneck over $h_t^c$. Its output, the FACET token $z_t=E_z(h_t^c)$, lets local policy adaptation consume the aligned scene state without backpropagating into the backbone.

\textbf{Actor objective.} The learner is a TD3-style deterministic actor--critic \cite{fujimoto2018td3} with twin critics and target networks. The critics are defined on the executable Cartesian action and regress onto the command produced by $\mathcal{S}_{\mathrm{task}}$. Successful human interventions provide the action anchor \cite{fujimoto2021td3bc,luo2024hilserl}, while realized next-step wrench supervises the auxiliary consequence head:
\begin{equation}
\begin{aligned}
\mathcal{L}_{\mathrm{actor}}(\eta) ={}&
-\mathbb{E}_{\mathrm{on}}\!\left[
  Q_{\xi_1}(e_t,a_t)
  \right] \\
&+\lambda_{\mathrm{BC}}\,
  \mathbb{E}_{\mathrm{BC}}\!\left[
  \left\|M_{\mathcal{T}}\odot
  (\bar a_t-\bar a_t^{\mathrm{human}})\right\|_2^2
  \right] \\
&+\lambda_{\mathrm{pred}}\,
  \mathbb{E}_{\mathrm{on}}\!\left[
  \mathcal{L}_w(\widehat w_{t+1},w_{t+1})
  \right].
\end{aligned}
\label{eq:actorloss}
\end{equation}
Here $\mathbb E_{\mathrm{on}}$ and $\mathbb E_{\mathrm{BC}}$ average over online transitions and successful intervention samples, respectively; $\bar a_t$ and $\bar a_t^{\mathrm{human}}$ are the normalized actor and intervention actions; $M_{\mathcal{T}}$ masks dimensions outside the task action space; and $\lambda_{\mathrm{BC}}$ and $\lambda_{\mathrm{pred}}$ weight action anchoring and wrench prediction.

\textbf{Deployment interface.} The deployed actor retains both heads, but only $e_t \to f_\eta \to a_t \to \mathcal{S}_{\mathrm{task}} \to$ robot is executable. The actor input concatenates the FACET token $z_t$, the normalized proprioceptive vector $[\,x_t; g_t; w_t\,]$ of Eq.~\eqref{eq:state}, and the frozen reference $\widetilde a_t$. The wrench head remains fully supervised, predictive, and non-commanded.

\textbf{Feature construction.} The four modality embeddings are
\begin{equation*}
e^{z} = \mathrm{LN}(W_z z_t), \quad
e^{p} = \mathrm{LN}(W_p [x_t,g_t]), \quad
e^{w} = \mathrm{LN}(W_w w_t), \quad
e^{\mathrm{ref}} = \mathrm{LN}(W_a \widetilde a_t),
\end{equation*}
where $\mathrm{LN}$ is layer normalization and $W_z$, $W_p$, $W_w$, and $W_a$ are learned modality projections. Each embedding is in $\mathbb{R}^{256}$, and their concatenation forms the actor input $e_t \in \mathbb{R}^{1024}$ of Sec.~\ref{sec:local_policy_adaptation}. A shared MLP actor $f_\eta$ produces $[u^a_t,\widehat w_{t+1}]$: the raw action activation $u^a_t\in\mathbb{R}^{7}$ is squashed into the per-task box $[a_{\min},a_{\max}]$ to form $a_t$, and $\widehat w_{t+1}\in\mathbb{R}^{6}$ is supervised by the measured next-step wrench through Eq.~\eqref{eq:actorloss}. The twin critics consume only $a_t$ as the executable action and do not treat $\widehat w_{t+1}$ as an action dimension.

\textbf{Why the critic scores the applied action.} $\mathcal{S}_{\mathrm{task}}$ can alter $a_t$ before execution, so replay stores the safety-filtered value in the same action field. This convention keeps the critic's regression target consistent with the observed dynamics without introducing a second action symbol.

\textbf{Why behavior cloning is success-filtered.} An intervention records what a human did, not that it worked, so promoting every intervention would anchor the actor to recoveries that themselves failed. Filtering on later confirmation keeps only segments known to reach the sub-goal, which matters most early in adaptation when the RL term is still driven by sparse, noisy reward. The confirmation key never enters the observation, so the policy cannot learn to read it.

\textbf{Sample-efficiency details.} Two choices support on-robot sample efficiency. \emph{Single-step commands}: the actor emits one absolute Cartesian target per decision rather than a chunk of horizon $H$, so every environment step yields a transition and the credit assignment horizon stays short. \emph{Bounded output}: the $\tanh$ squashing into per-task limits $[a_{\min},a_{\max}]$ (Sec.~\ref{sec:local_policy_adaptation}) replaces the small-correction assumption that residual baselines rely on: the actor may propose targets far from the reference, but never outside the task's admissible box.

\textbf{Representation-to-policy interface.} The FACET token and the measured wrench play complementary roles at the local actor. The token carries task intent and the contact structure learned from the larger corpus, whereas $w_t$ reports the interaction currently unfolding on the new part. Conditioning on both allows the same aligned scene state to support different bounded targets when measured contact differs. The frozen reference supplies the nominal task motion, while the actor learns only the part-dependent correction expressed as an absolute target.

\begin{algorithm}[H]
\caption{Bounded local policy adaptation with an auxiliary wrench head}
\label{alg:local_policy_adaptation}
\begin{algorithmic}[1]
\Require frozen $\{\pi_{\mathrm{ref}},E_z,\Omega\}$; actor $f_\eta$; twin action critics $Q_{\xi_i}$; bounds $a_{\min},a_{\max}$; mask $M_{\mathcal{T}}$; operator $\mathcal{S}_{\mathrm{task}}$
\Ensure bounded action policy and supervised auxiliary wrench predictor
\Loop
  \State $z_t \leftarrow E_z(h_t^c)$ \Comment{FACET token from the aligned representation}
  \State $\widetilde a_t \leftarrow$ reference action from frozen $\pi_{\mathrm{ref}}$;\quad $e_t \leftarrow [\,z_t;\, \mathrm{Norm}(s^{\mathrm{kin}}_t, w_t);\, \widetilde a_t\,]$ \Comment{conditioning, not a base}
  \State $\left[u^a_t,\widehat w_{t+1}\right] \leftarrow f_\eta(e_t)$;\quad $a_t \leftarrow a_{\min} + \tfrac{1}{2}(\tanh u^a_t + 1) \odot (a_{\max}-a_{\min})$ \Comment{Sec.~\ref{sec:local_policy_adaptation}}
  \State $a_t \leftarrow \mathcal{S}_{\mathrm{task}}(a_t)$;\quad execute $a_t$ \Comment{wrench prediction is non-commanded}
  \State $\mathcal{D}_{\mathrm{online}} \leftarrow \mathcal{D}_{\mathrm{online}} \cup \{(e_t,a_t,r_t,d_t,e_{t+1},w_{t+1})\}$
  \If{human intervened and the segment is later confirmed successful}
    \State $\mathcal{D}_{\mathrm{BC}} \leftarrow \mathcal{D}_{\mathrm{BC}} \cup \{(e_t, \widetilde a_t, a_{\mathrm{human},t})\}$
  \EndIf
  \State $\hat q_t \leftarrow r_t + \gamma(1-d_t)\min_i Q_{\bar\xi_i}\big(e_{t+1},\pi^a_{\bar\eta}(e_{t+1})\big)$
  \State $\xi_i \leftarrow \xi_i - \alpha_Q \nabla_{\xi_i}\mathbb{E}\big[(Q_{\xi_i}(e_t,a_t)-\hat q_t)^2\big]$
  \State $\eta \leftarrow \eta - \alpha_\pi \nabla_\eta \mathcal{L}_{\mathrm{actor}}$ \Comment{Eq.~\eqref{eq:actorloss}}
\EndLoop
\end{algorithmic}
\end{algorithm}
Here $\bar\xi_i$ and $\bar\eta$ denote target-critic and target-actor parameters, $\pi^a_{\bar\eta}$ is the action head of the target actor, $\hat q_t$ is its one-step TD target, and $\alpha_Q$ and $\alpha_\pi$ are the critic and actor learning rates.

\textbf{Separation of reference, command, and consequence.} Algorithm~\ref{alg:local_policy_adaptation} keeps three quantities distinct. The frozen action $\widetilde a_t$ is a reference that informs the actor but is never added as a residual base. The bounded output $a_t$ is the proposed Cartesian target, and the value stored in replay is the target after the task safety operator has been applied. The auxiliary $\widehat w_{t+1}$ predicts the measured consequence of that applied motion. This separation lets the critic remain a value function over the executable action space while the shared actor trunk retains an explicit model of how a correction changes contact.

\textbf{Frozen and adaptable information.} The semantic--contact encoder, FACET bottleneck, reference policy, VLM judge, compliant controller, and safety operator remain fixed. Only the lightweight actor and its twin critics adapt to the new part. The boundary is intentional: language semantics, scene interpretation, and coarse task progress are reusable across parts, whereas local mass, stiffness, latch response, and contact geometry determine which bounded correction succeeds. Freezing the upstream representation prevents a small on-robot dataset from rewriting those reusable quantities, while conditioning on the reference keeps the adapted actor anchored to the task-level behavior learned from the larger corpus.

\textbf{Asymmetric use of the two actor heads.} Both heads remain active throughout adaptation, but they enter optimization differently. The action head is evaluated by TD3 and is the only head connected to execution. The wrench head is trained only against future wrist F/T and regularizes the shared trunk toward action--contact consistency. It neither enlarges the Markov decision process action nor closes the robot's force loop; measured feedback remains the input to compliant control. The local update therefore changes how the policy reacts to contact without introducing a second force-command interface or changing the safety contract used by the rest of the system.

\textbf{Consistency under safety filtering.} The transition stored in replay contains the command that actually reaches the controller, not the actor's pre-filter proposal. This distinction matters whenever workspace clipping or rate limits alter a target: the observed reward and next state were caused by the applied command, so regressing the critic on the unapplied proposal would assign value to an action the robot never took. The actor can still improve its bounded proposal through the critic, but every target used for value learning remains grounded in the closed-loop trajectory. The safety operator therefore stays external to the learned policy without creating an inconsistency between optimization and execution.

\clearpage

\section{Reported Per-Sub-Goal Results}
\label{app:results}

Table~\ref{tab:main} pairs the five task-level results of Table~\ref{tab:task_success} with a reported subset of 17 of the suite's 23 sub-goals, including 13 contact-critical rows. The subset is visualized for representative methods in Fig.~\ref{fig:main_results} and diagnoses whether failures concentrate at contact; it is not presented as a complete 23-row decomposition. Aggregate contact means are unweighted over the 13 shown contact rows and rounded to the nearest point. The \emph{Align}, \emph{+RL}, and \emph{Full} columns match the variant names used in the body.

\begingroup
\begin{table}[H]
\centering
\caption{Reported task and sub-goal success rates (\%), 20 trials per cell. The 17 displayed sub-goals are a diagnostic subset of the 23-sub-goal suite; $\dagger$ marks the 13 contact-critical rows. Shaded rows report task-level chain success.}
\label{tab:main}
\setlength{\tabcolsep}{3.0pt}
\renewcommand{\arraystretch}{1.04}
\small
\begin{tabular*}{\linewidth}{@{\extracolsep{\fill}}lcccccccc@{}}
\toprule
& \multicolumn{5}{c}{Baselines} & \multicolumn{3}{c@{}}{\system{} (ours)} \\
\cmidrule(lr){2-6}\cmidrule(l){7-9}
Sub-goal & $\pi_{0.5}$ & +F & +RECAP & GR00T & TA-VLA & Align & +RL & \textbf{Full} \\
\midrule
\multicolumn{9}{@{}l}{\textit{RAM installation}} \\
\quad pick & 100 & 100 & 100 & 95 & 100 & 100 & 100 & 100 \\
\quad align$^\dagger$ & 10 & 15 & 35 & 10 & 10 & 15 & 45 & 95 \\
\quad place$^\dagger$ & 40 & 65 & 70 & 40 & 60 & 65 & 75 & 100 \\
\quad press$^\dagger$ & 80 & 85 & 95 & 75 & 80 & 85 & 100 & 95 \\
\rowcolor{gray!10}
\quad \textbf{RAM task} & 10 & 15 & 35 & 10 & 10 & 15 & 45 & \textbf{95} \\
\midrule
\multicolumn{9}{@{}l}{\textit{CPU installation}} \\
\quad pick & 80 & 80 & 100 & 80 & 100 & 100 & 100 & 100 \\
\quad align$^\dagger$ & 5 & 5 & 15 & 5 & 20 & 20 & 45 & 85 \\
\quad place$^\dagger$ & 55 & 60 & 65 & 55 & 60 & 65 & 75 & 95 \\
\quad press$^\dagger$ & 70 & 70 & 70 & 65 & 70 & 75 & 80 & 95 \\
\rowcolor{gray!10}
\quad \textbf{CPU task} & 5 & 5 & 15 & 5 & 20 & 20 & 45 & \textbf{85} \\
\midrule
\multicolumn{9}{@{}l}{\textit{Disk installation}} \\
\quad pick & 65 & 70 & 65 & 60 & 90 & 100 & 100 & 100 \\
\quad align$^\dagger$ & 25 & 20 & 20 & 5 & 30 & 30 & 65 & 95 \\
\quad place$^\dagger$ & 70 & 70 & 75 & 75 & 90 & 90 & 100 & 100 \\
\rowcolor{gray!10}
\quad \textbf{Disk task} & 25 & 20 & 20 & 5 & 30 & 30 & 65 & \textbf{95} \\
\midrule
\multicolumn{9}{@{}l}{\textit{GPU installation}} \\
\quad pick & 70 & 70 & 70 & 75 & 80 & 100 & 100 & 100 \\
\quad align$^\dagger$ & 10 & 5 & 5 & 0 & 5 & 10 & 35 & 85\\
\quad place$^\dagger$ & 25 & 25 & 30 & 25 & 25 & 30 & 85 & 95 \\
\quad press$^\dagger$ & 20 & 20 & 25 & 25 & 60 & 60 & 65 & 90 \\
\rowcolor{gray!10}
\quad \textbf{GPU task} & 10 & 5 & 5 & 0 & 5 & 10 & 35 & \textbf{85} \\
\midrule
\multicolumn{9}{@{}l}{\textit{LEVER closure}} \\
\quad press$^\dagger$ (first) & 0 & 0 & 0& 0 & 5 & 5 &0 & 50 \\
\quad press$^\dagger$ (second) & 0 & 0 & 0& 0 & 5 & 5 &0 & 55 \\
\rowcolor{gray!10}
\quad \textbf{LEVER task} & 0 & 0 & 0 & 0 & 5 & 5 & 0 & \textbf{50} \\
\midrule
Mean, 13 reported contact sub-goals$^\dagger$ & 32 & 34 & 39 & 29 & 40 & 43 & 59 & \textbf{87} \\
\rowcolor{gray!15}
\textbf{Mean, five tasks} & 10 & 9 & 15 & 4 & 14 & 16 & 38 & \textbf{82} \\
\bottomrule
\end{tabular*}
\end{table}
\endgroup

\textbf{Reading the decomposition.} Task rows require completion of the full chain, whereas sub-goal rows isolate individual decision points. For the full system, each chain rate matches its weakest displayed contact-critical sub-goal: alignment limits RAM, CPU, Disk, and GPU, while the first press limits LEVER. The 13-row contact mean therefore measures local interaction quality with equal weight per event; the five-task mean measures end-to-end reliability with equal weight per assembly. Reporting both separates contact quality from chain completion without allowing longer tasks to dominate the diagnostic average.

\clearpage

\section{Supporting Evaluation Data}
\label{app:supporting_results}

This section retains the exact reference values behind the compact visualizations in the main text, together with the deployment-properties table moved out of the main narrative. On the same fixtures, \system{} completes the task 40\% faster than expert teleoperation.

\begingroup
\setlength{\intextsep}{7pt}
\begin{table}[H]
\centering
\caption{Deployment efficiency and precision; placement accuracy is measured at the align-to-insert transition.}
\label{tab:efficiency}
\setlength{\tabcolsep}{6pt}
\renewcommand{\arraystretch}{1.12}
\small
\begin{tabular}{@{}lcc@{}}
\toprule
Method & Inference latency (ms) $\downarrow$ & Placement accuracy (mm) $\downarrow$ \\
\midrule
$\pi_{0.5}$ \cite{intelligence2025pi05} & 150 & $\approx$5 \\
\rowcolor{gray!10}
\textbf{\system{} (Full)} & \textbf{50} & \textbf{0.5} \\
\bottomrule
\end{tabular}
\end{table}

\begin{table}[H]
\centering
\caption{Disk post-training results underlying Fig.~\ref{fig:posttrain_bar}, with matched deployment data, VLM reward, and budget.}
\label{tab:posttrain}
\setlength{\tabcolsep}{8pt}
\renewcommand{\arraystretch}{1.12}
\small
\begin{tabular}{@{}lcc@{}}
\toprule
Metric & Matched AWR & \textbf{\system{} post-training} \\
\midrule
Success rate $\uparrow$ & 20\% & \textbf{65\%} \\
Intervention rate $\downarrow$ & 47\% & \textbf{24\%} \\
Recovery rate $\uparrow$ & 44\% & \textbf{81\%} \\
\bottomrule
\end{tabular}
\end{table}

\begin{table}[H]
\centering
\caption{Adaptation reference values for Fig.~\ref{fig:adaptation}. Results use a two-hour window and a 30-min checkpoint; violations cover all training episodes, and -- means 90\% success was not reached.}
\label{tab:adapt_paradigm}
\setlength{\tabcolsep}{5pt}
\renewcommand{\arraystretch}{1.15}
\footnotesize
\begin{tabular}{@{}lccc@{}}
\toprule
Method & To 90\% (min) $\downarrow$ & Success @ 30\,min (\%) $\uparrow$ & Violations (\%) $\downarrow$ \\
\midrule
DSRL \cite{wagenmaker2025dsrl} & -- & 10 & 90 \\
Residual off-policy \cite{residualoffpolicy2025} & 35.1 & 80 & 20 \\
\rowcolor{gray!10}
\textbf{\system{} local adaptation} & \textbf{10.5} & \textbf{95} & \textbf{5} \\
\bottomrule
\end{tabular}
\end{table}
\endgroup

\subsection{Behavioral Analysis}
\label{app:behavior}

This subsection records the mechanisms underlying the takeaways of Sec.~\ref{sec:exp}. The body states the conclusion; the supporting detail is kept here.

\textbf{Contact-aware retry.} The largest margins over $\pi_{0.5}$ occur on RAM and CPU, where a visually aligned insertion can still meet the slot rim. The base policy continues the already issued downward chunk, whereas the adapted actor uses the resulting off-axis wrench to retreat, correct laterally, and re-enter (Fig.~\ref{fig:adaptation}). The correction remains within the same sub-goal and runs at the contact-control rate, so it does not require the backbone to re-plan the task.

\textbf{Early failure observability.} During a press-and-lock sub-task, gripper occlusion can make a slight misseat look identical to successful seating. Wrist F/T separates the two cases: force rises while displacement stalls, allowing the system to flag the jam, re-issue the sub-goal, and complete the insertion. This contact-specific observability underlies the recovery gain in Sec.~\ref{sec:exp_posttrain}.

\textbf{Protected recovery credit.} Because Eq.~\eqref{eq:reward} is evaluated per sub-task, a recovered trajectory receives credit for its corrective segment rather than being discarded with the preceding failure. Regime-conditioned selection then keeps these rare contact-recovery frames in the positive set. Together, the two mechanisms improve the tail behavior that determines whether a failed interaction resumes or halts the cell.

\textbf{Separation of evaluation scales.} Table~\ref{tab:main} measures complete chains and individual contact decisions, Table~\ref{tab:posttrain} isolates changes from deployment post-training, and Table~\ref{tab:adapt_paradigm} records the speed and safety of local adaptation. These quantities retain separate denominators because they describe different operating scales; reading them together connects task completion to recovery and adaptation without collapsing distinct outcomes into a single aggregate score.

\FloatBarrier

\end{document}